\documentclass[12pt, oneside]{article}
\usepackage[T1]{fontenc}
\usepackage[utf8]{inputenc}

\usepackage[english]{babel}
\usepackage{mathrsfs}
\usepackage{mathtools}
\usepackage{bm}
\usepackage{booktabs}
\usepackage[format=plain,width=0.9\textwidth,font=small,labelfont=bf]{caption}
\usepackage{multirow}
\usepackage[pagewise]{lineno}\linenumbers
\modulolinenumbers[5]
\usepackage{rotating}
\usepackage{subcaption}
\usepackage{adjustbox}
\usepackage{caption}

\numberwithin{figure}{section}
\numberwithin{table}{section}
\usepackage{amsbsy}
\usepackage{amstext}
\usepackage{amsthm}
\usepackage{amssymb} 
\usepackage{setspace}
\usepackage[sort,authoryear,round]{natbib}
\usepackage{xcolor}
\usepackage{hyperref}
\hypersetup{%
  colorlinks=true,
  citebordercolor={0 1 0}
}
\usepackage{float}
\makeatletter

\providecommand{\tabularnewline}{\\}

\numberwithin{equation}{section}
\numberwithin{figure}{section}

\usepackage{adjustbox}
\usepackage{enumitem}
\usepackage{placeins}
\setlist[itemize]{noitemsep, topsep=0pt}
\usepackage{url}
\usepackage{color}
\usepackage{authblk}
\usepackage{fullpage}
\usepackage[ruled,linesnumbered]{algorithm2e}
\newlength\mylen

\let\oldnl\nl% Store \nl in \oldnl
\newcommand{\nonl}{\renewcommand{\nl}{\let\nl\oldnl}}% Remove line number for one line

\newcounter{algoline}
\newcommand\Numberline{\refstepcounter{algoline}\nlset{\thealgoline}}
\DeclareMathOperator{\RR}{\mathbb{R}}

\newtheorem{theorem}{Theorem}
\def\spacingset#1{\renewcommand{\baselinestretch}%
{#1}\small\normalsize} \spacingset{1}

\makeatother
\definecolor{spirodiscoball}{rgb}{0.06, 0.75, 0.99}

\begin{document}
%\singlespacing
\onehalfspacing
%\doublespacing
\title{Spherical Rotation Dimension Reduction with Geometric Loss Functions}
\author[1]{\small Hengrui Luo}
\author [2] {\small Jeremy E. Purvis}
\author[3]{\small Didong Li}
\affil[1]{\footnotesize Lawrence Berkeley National Laboratory, Berkeley, CA, 94720, USA, E-mail: hrluo@lbl.gov}
\affil[2]{\footnotesize Department of Genetics, University of North Carolina at Chapel Hill, Chapel Hill, NC, 27599, USA, E-mail: purvisj@email.unc.edu}
\affil[3]{\footnotesize Department of Biostatistics, University of North Carolina at Chapel Hill, Chapel Hill, NC, 27599, USA, E-mail: didongli@unc.edu}
%\author{Hengrui Luo and Didong Li}
\date{}%\date{\centering{\today}\vspace{-3cm}}

\maketitle
\begin{abstract}

Modern datasets often exhibit high dimensionality, yet the data reside in low-dimensional manifolds that can reveal underlying geometric structures critical for data analysis. A prime example of such a dataset is a collection of cell cycle measurements, where the inherently cyclical nature of the process can be represented as a circle or sphere. Motivated by the need to analyze these types of datasets, we propose a nonlinear dimension reduction method, Spherical Rotation Component Analysis (SRCA), that incorporates geometric information to better approximate low-dimensional manifolds. SRCA is a versatile method designed to work in both high-dimensional and small sample size settings. By employing spheres or ellipsoids, SRCA provides a low-rank spherical representation of the data with general theoretic guarantees, effectively retaining the geometric structure of the dataset during dimensionality reduction. A comprehensive simulation study, along with a successful application to human cell cycle data, further highlights the advantages of SRCA compared to state-of-the-art alternatives, demonstrating its superior performance in approximating the manifold while preserving inherent geometric structures.

\end{abstract}

\noindent%
{\it Keywords: }  principal component analysis, high-dimensional dataset, dimension reduction.

\section{Introduction}
Modern data analysis presents the challenge of high-dimensionality, where the dataset usually comes as high dimensional vectors in $\mathbb{R}^d$, with a large $d$. 
Dimension reduction (DR) methods seek low dimensional representation of high dimension data \citep{mukhopadhyay2020estimating,zhang2021animal} to facilitate data visualization, subsequent data exploration, and statistical modeling in machine learning \citep{jolliffe2016principal}. Along with the difficulty in visualizations and computation, non-linearity obstructs conventional dimension reduction methods.
\subsection{Motivation: Human Cell Cycle}

In traditional DR methods (e.g., Principal Component Analysis (PCA), \citet{pearson1901liii}), it has been repeatedly pointed out that normalization preprocessing, including translations (by mean) and scalings (by standard deviation), is crucial in practicing DR \citep{jolliffe1995rotation}. However, rotation as a preprocessing step is less studied in the DR context. We are motivated by preserving non-trivial geometrical structure in DR tasks, and observed that \emph{rotations} are as important as translations and scalings if we want to design DR methods that respect the underlying structure.

A compelling example that illustrates the need for advanced dimension reduction methods respecting the underlying structure is the analysis of cell cycle data. The cell cycle is an inherently cyclical process \citep{schafer1998cell} that consists of four proliferative phases: G1, S, G2, and M. Fluctuations in cell cycle genes and proteins show periodic, non-linear trends, that can be represented as a circle or sphere in a lower-dimensional space. Traditional linear methods may not adequately capture these properties, leading to the loss of crucial information.

\begin{figure}[h!]
	\centering\includegraphics[width=0.9\textwidth]{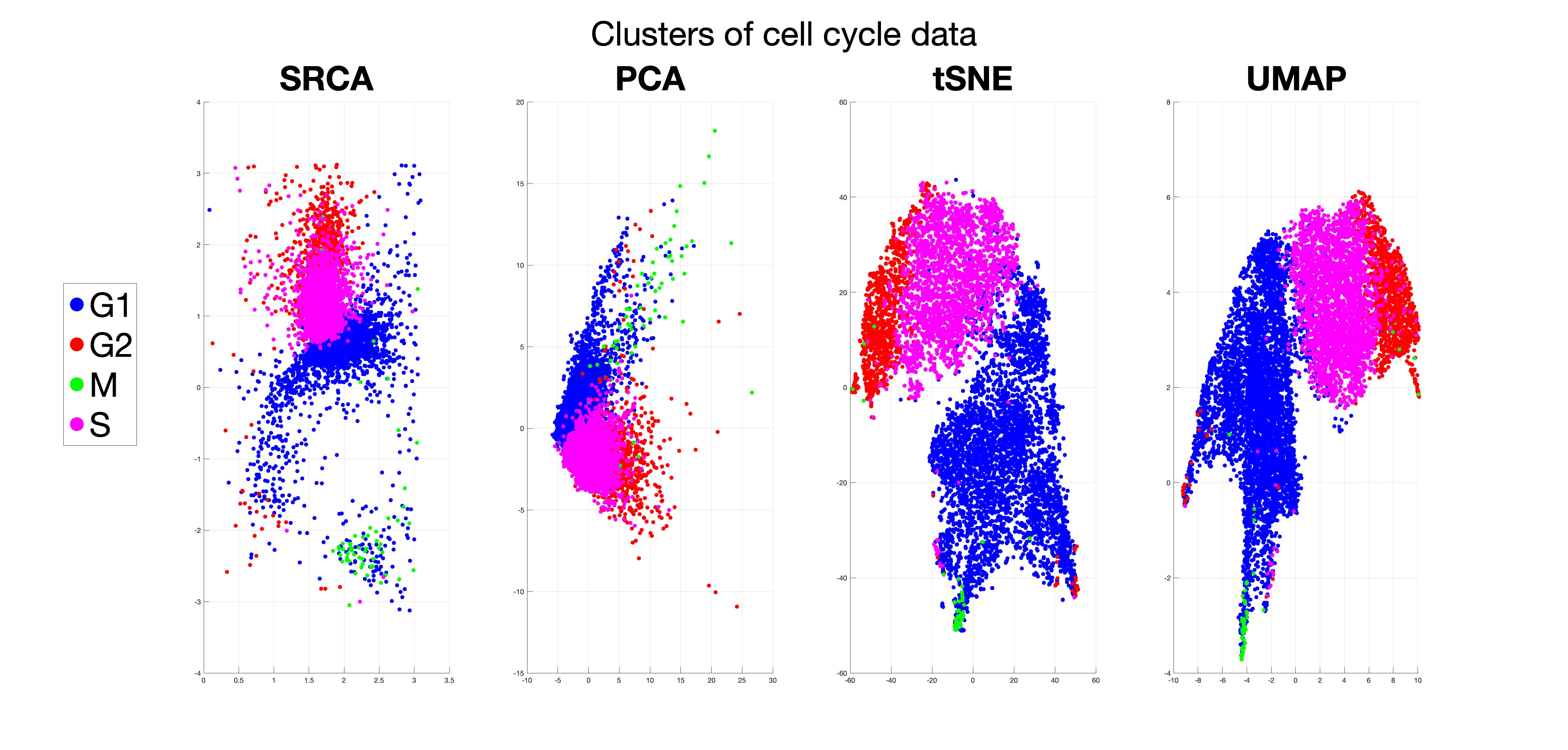}
\caption{2-dimensional representation of cell cycle data, colored by different cell phases.}
\label{fig:motivation}
\end{figure}

Figure \ref{fig:motivation} presents a 2-dimensional representation of cell cycle data proposed in \cite{stallaert2022structure}, which included 40 single-cell features such as the expression or localization of core cell cycle regulators and signaling proteins. These features combine to form a multivariate cell cycle signature for each cell in the entire population, collected from 8,850 individual cells. Because individual cells are naturally asynchronous during data collection, the cells are randomly sampled over the entire cyclical distribution of possible cell cycle states. The phase of each cell (G1, S, G2, or M) was assigned using its unique molecular profile. Based on the known sequence of cell cycle phases, we would expect consecutive phases, such as G2 (red) and M (green) to be neighbors in the low dimensional projection. However, existing methods such as PCA, t-distributed Stochastic Neighbor Embedding (tSNE, \cite{van2008visualizing}), and Uniform Manifold Approximation and Projection (UMAP, \cite{mcinnes2018umap-software}) (selected from the best results among other methods attempted) fail to preserve this structure in their representations.

This example motivates the development of a new DR method that utilizes spheres to represent high-dimensional data in low-dimensional spaces, effectively preserving the geometric structure and inherent cyclical nature of biological processes. In contrast to other DR methods, our proposed method, provides a representation on a 2-dimensional sphere, represented by longitude and latitude in the first panel (see Section \ref{sec:cell_cycle} for more details). This SRCA representation in the lower dimensional space clearly preserves the cell cycle progression: G1 $\rightarrow$ S $\rightarrow$ G2 $\rightarrow$ M $\rightarrow$ G1, where the latitude (y-axis) is in the $\mathrm{mod}~2\pi$ sense, meaning $\pi=-\pi$. This biological periodicity, or in other words sphericity, is of central importance in analyzing cell data \citep{stallaert2022structure}. % (See Appendix \ref{sec: MSE comparison} for synthetic).
In general, disruption of these low-dimensional structures in a high-dimensional dataset \citep{luo_asymptotics_2020, luo2022nonparametric} diminishes the effectiveness of the subsequent analysis procedure like clustering and classification.

\begin{table}[h!]
	\begin{centering}
		\begin{tabular}{|c|c|c|}
			\hline 
		    & Local & Global\tabularnewline
			\hline 
			Manifold learning  & LLE, tSNE, UMAP & MDS, Isomap, GPLVM \tabularnewline
			\hline 
			Manifold estimation & PCurv, GMRA, LPE, SAME, Spherelets & PCAs, SPCA, SRCA\tabularnewline
			\hline 
		\end{tabular}
		\par\end{centering}
	\caption{Conceptual categorization of selected dimension reduction methods}
	\label{table:DR} 
\end{table}

\subsection{Related Literature}
Sphericity induced by periodicity in the above data example requires the development of sophisticated non-linear DR methods designed to preserve certain structures in the data. The common assumption is that the observetaions $x_1,\cdots,x_n$ are near a manifold $M$ embedded in $\RR^d$. Table \ref{table:DR} provides a selected collection of dimension reductions methods loosely categorized in two ways. Algorithms in the first row are known as ``manifold learning'' \citep{lin2008riemannian}, which output some low dimensional features in a new Euclidean space  of dimension $d'$ instead of an estimate of $M$, denoted by $\widehat{M}$. These methods include Locally Linear Embedding (LLE, \cite{roweis2000nonlinear}),  tSNE, , UMAP, Multi-Dimensional Scaling (MDS, \cite{kruskal1978multidimensional}),  Isomap \citep{tenenbaum2000global}, Gaussian Processes Latent Variable Model (GPLVM, \cite{titsias2010bayesian}), etc. 

%As a consequence, the performance of the DR method, usually presented by the mean squared distance between observation points $x_i$ and their projections $\widehat{x}_i$ on the estimated manifold, is not well-defined. 
%The mean squared distance $\frac{1}{n}\sum_{i=1}^n d^2(x_i,\widehat{x}_i)$  is  a straightforward metric that can be used to define a DR method, but allows cross-validation for tuning parameters in DR methods. For example, in PCA, the mean squared distance, also known as the geometric loss, is used to determine the number of principal components and hence the dimension of the low dimensional linear subspace. 

In contrast, the other type of DR methods, known as ``manifold estimation'', which estimates $M$ in $\RR^d$ directly, has been attracting researchers' attention \citep{genovese2012minimax}. There is an immense literature in local methods including Principal Curves (PCurv), Geometric Multi-Resolution Analysis (GMRA~\cite{allard2012multi}), Local Polynomial Estimator (\cite{aamari2019nonasymptotic}), Structure-Adaptive Manifold Estimation (SAME~\cite{puchkin2022structure}), Spherelets~\citep{li_efficient_2022}, etc. The common idea behind these methods is to partition the space into local regions, and apply local, often linear, method to each small region. The intuition is that a manifold can be locally approximated by its tangent spaces. However, these local, nonparametric, complex methods are often computationally expensive and lack of interpretability.

%The objective of SPCA is not to minimize the geometric loss defined by the mean square distance. Instead, an algebraic loss based algorithm is adopted for a closed-form solution of SPCA. 

A recent attempt to develop a spherical analogue of PCA is \citet{li_efficient_2022}, which allows us to conduct dimension reduction and learn the shape of spherically distributed datasets. However, both PCA and SPCA fail when the sample size $n<d'$, the retained dimension (i.e., the dimension of the reduced dataset, the formal definition is introduced below) and are not %\st{scalable}
easily applicable to high dimensional datasets. For instance, in the gene expression data, $d$ is the number of genes, often over $20,000$ and the retained dimension $d'$ is often chosen to be a couple of hundreds with the largest variability \citep{townes2019feature}. While the sample size could be much smaller, for example, less than $20$ for certain tissues in the Genotype-Tissue Expression (GTEx) dataset \citep{gtex2020gtex}. In fact, most existing dimension reduction methods cannot handle $n<d'$ without substantial modifications. 

%In the current paper, we explore a geometrically induced loss function with a fully statistically principled dimension reduction approach called Spherical Rotation Component Analysis (SRCA) that is applicable to datasets with large original dimension $d$, large retained dimension $d'$ and the situation where  $n<d'$.
In this paper, we focus on parametric global methods  
%There has been plenty of attempts to extend PCA in different ways, like kernel PCA, sparse PCA, generalized PCA, etc. While, for dataset with sphericity, a recent attempt to develop a spherical analogue of PCA is \citet{li_efficient_2022}, which allows us to conduct dimension reduction and learn the shape of spherically distributed datasets. SPCA works well for datasets that live on a low-dimensional sphere. Intuitively, this problem
%is well-defined as an estimation problem of the sphere paramerized by its center and radius. The geometric constraints imposed by the SPCA method help us identify low-dimensional structures in the reduced dataset, which turns out to be extremely challenging
%\citep{luo_asymptotics_2020, luo2022nonparametric}. 
and derive an DR method called \emph{spherical rotation dimension reduction} (SRCA), that preserves the sphericity constraints of the dataset. Unlike some competitors, this method is applicable to high-dimensional datasets regardless of retained dimensions and sample sizes. %Before formal expositions, we use toy examples to illustrate these points. The principal component analysis (PCA, \cite{pearson1901liii}) is linear in the sense that the dimension-reduced data $\widehat{x}_i$ can be obtained
%by linear transformations on the observed data $x_i\in\RR^d$. 
SRCA is scalable and interpretable, and will not destroy not only the geometry but also the 
topology of the dataset.  (see Supplement \ref{sec:orthogonalLoop} for synthetic examples).

Specifically, we focus on biological and genetic datasets, where dimension reduction is adopted by biologist directly before clustering and subsequent tasks \citep{johnson2022embedr, zhou2018using}. 
Our method also echoes and exemplifies a grander community belief that any dimension reduction should be guided by the need of its subsequent analyses and respect the structure in the original dataset.

\section{Methodology}\label{sec: methodology}
In this section, we outline the proposed procedure that aims to minimize a geometric loss function, specifically the mean squared errors between the original data points $x_i$ and dimension-reduced data points $\widehat{x_i}$. %There are various definitions for the dimensionality of the reduced space, and for the remainder of the paper, we will use the following concept of retained dimension.

%\noindent \textbf{Retained dimension.\label{RemarkOnRetainedDim}} In the subsequent discussions, 
We denote the intrinsic dimension of the support of the reduced dataset by $d'$ and refer to it as the \emph{retained dimension}. It is worth noting that some literature uses the embedded dimension as $d'$. For example, if the reduced dataset lies in $\mathbb{S}^1$ embedded into $\mathbb{R}^2$, we would consider the dimensionality of the reduced data to be $d'=1$ and not $2$, as $\mathbb{S}^1$ is a one-dimensional manifold.

\subsection{PCA and SPCA Revisited}\label{sec: PCA and SPCA Revisited}

We begin by recalling that PCA identifies a low-rank linear subspace from observations $\mathcal{X}={x_{1},\cdots,x_{n}}\subset\RR^d$ by minimizing the sum of squared error loss function: 
\begin{align*}
	\min_{{V}\in\mathbb{R}^{d\times d'}} \sum_{i=1}^n\|{x}_i-\widehat{{x}_i}\|^2 = \sum_{i=1}^n\|{x}_i-{\bar{x}}-{V}{V}^{T}({x}_i-{\bar{x}})\|^2,	%\sum{}_{i=1}^{n}({x}_{i}-{\bar{x}})^{T}{U}^{T}{U}({x}_{i}-{\bar{x}}), 
	\text{s.t. }{V}^{T}{V}={I_{d'}}.
\end{align*}
where ${\bar{x}}=\frac{1}{n}\sum_{i=1}^{n}{x}_i$ is the sample mean calculated in $\mathbb{R}^d$. The solution to this optimization problem yields a rotation matrix $V$ that defines a subspace, called the solution to PCA.

From an optimization perspective, PCA is a problem for a given (geometric) loss function \citep{journee2010generalized}, which quantifies the $l_2$ errors between the observation and the subspace. SPCA \citep{li_efficient_2022} aims to find the optimal sphere $S$. Unlike PCA's planar solution, the solution of SPCA is a sphere with center $c$ and radius $r$ residing in the linear subspace $V$ to fit the data. SPCA does not minimize the sum of squared distances between observations and the sphere; instead, it employs a two-step algorithm to minimize the sum of point-to-plane and projection-to-sphere distances.

The desired one-step algorithm was not explored in the original paper \citep{li_efficient_2022} since the problem is theoretically more complicated and lacks a closed-form solution (See Supplement \ref{sec:Sphere-Estimation-with}). The SRCA can be shown to attain this one-step goal. %In this paper, we revisit the loss function and develop a geometric loss-based non-linear DR method that allows for one-step optimization, estimating the spherical space and dimensionality simultaneously. A significant advantage of this geometric loss-based method is its applicability to cases with large $d$ and $n<d'$.

\subsection{Geometric Loss Function}\label{sec: geometric loss function}

Given a sphere centered at ${c}$ with radius $r$, we first assume that
it lies in a subspace parallel to a coordinate plane in $\mathbb{R}^{d}$ determined by $\mathcal{I}\subset\{1,\cdots,d\}$ after a linear transformation determined by a (non-singular) matrix ${W}$. 
We denote such low-dimensional ``sub-sphere'' by $S_{\mathcal{I}}({c},r)$ and use the notation $I_{\mathcal{I}}$ to denote an identity matrix with ones in $(i,i)$-th entries, $i\in\mathcal{I}$ but zeros in the rest entries, then the point-to-sphere distance from a generic point $x_i$ to this sphere can be expressed as
\begin{align*}
	d({x}_{i},S_{\mathcal{I}}({c},r))^{2} & =({x}_{i}-{c})^{T}{W}{I}_{\mathcal{I}^{c}}({x}_{i}-{c})+\left(\sqrt{({x}_{i}-{c})^{T}\sqrt{{W}}^{T}{I}_{\mathcal{I}}\sqrt{{W}}({x}_{i}-{c})}-r\right)^{2} \\
	& =({x}_{i}-{c}^{T}) {W}({x}_i-{c})+r^{2}-2r\sqrt{({x}_{i}-{c})^{T}\sqrt{{W}}^{T}{I}_{\mathcal{I}}\sqrt{{W}}({x}_{i}-{c})}. 
\end{align*}
Note that the assumption that the sphere lies in a coordinate plane is essential; otherwise no closed form expression can be written. With this point-to-sphere distance, the (geometric loss) function can be written as
\begin{align*}
	\mathscr{L}({c},r,\mathcal{I}\mid\mathcal{X},{W}) & = \sum_{i=1}^{n}\left(({x}_{i}-{c})^{T} {W}({x}_i-{c})+r^{2} 
	-2r\sqrt{({x}_{i}-{c})^{T}\sqrt{{W}}^{T}{I}_{\mathcal{I}}\sqrt{{W}}({x}_{i}-{c})}\right)\\
	& \eqqcolon \sum_{i=1}^{n}\rho(x_{i};\theta),
\end{align*}
where we use the notation $\theta=(c,r)$ with a given $\mathcal{I}$, and the notation $\rho(x_i;\theta)$ is adopted to emphasize its additive form and to facilitate the later theoretic discussions.
%With this notation, 
Our dimension
reduction procedure can be described as solving the optimization problem below: 
\begin{align}
	& \min_{\mathcal{I}\subset\{1,\cdots,d\},{c}\in\mathbb{R}^{d},r\in\mathbb{R}^{+}}\mathscr{L}({c},r,\mathcal{I}\mid\mathcal{X},{W})\label{eq:basic opt with c and r}  =\min_{\mathcal{I}\subset\{1,\cdots,d\},{c}\in\mathbb{R}^{d},r\in\mathbb{R}^{+}}\sum_{i=1}^{n}\left(({x}_{i}-{c})^{T} {W}({x}_i-{c})+r^{2}\right.\nonumber\\
	& \left.~~~ -2r\sqrt{\left({x}_{i}-{c}\right)^{T}\sqrt{{W}}^{T}{I}_{\mathcal{I}}\sqrt{{W}}\left({x}_{i}-{c}\right)}\right) \text{, s.t.}|\mathcal{I}|=d'+1
\end{align}
Using the loss function defined in (\ref{eq:basic opt with c and r}),
we can simultaneously estimate the center ${c}$, radius $r$ and $\mathcal{I}$ by solving the optimization problem.  Since there are
at most $2^{d}$ possible choices of $\mathcal{I}$, it is straightforward
to verify that this one-step optimization problem (\ref{eq:basic opt with c and r})
can be equivalently solved in a two-step procedure: First, select a subset of indices $\mathcal{I}\subset\{1,\cdots,d\}$, and then estimate the center ${c}$ and radius $r$ of the dataset $\mathcal{X}$. 
This optimization problem can also be solved iteratively on variables
$\mathcal{I},r,{c}$. The binary search can be the first step,
followed by esimating the center
${c}$ and radius $r$: 
\begin{enumerate}
	\item Given ${c}$ and $r$, perform an exhaustive binary search among all possible $\mathcal{I}$. 
	\item Given $\mathcal{I}$ and ${c}$, take the derivative of $\frac{\partial\mathscr{L}}{\partial r}=0$ to obtain:
	\[
	r=\frac{1}{n}\sum_{i=1}^{n}\sqrt{({x}_{i}-{c})^{T}\sqrt{{W}}^{T}{I}_{\mathcal{I}}\sqrt{{W}}({x}_{i}-{c})}.
	\]
	\item Given $\mathcal{I}$ and $r$, take the derivative of $\frac{\partial\mathscr{L}}{\partial {c}}={0}$ to obtain:
	\[
	\frac{\partial\mathscr{L}}{\partial{c}}=\sum_{i=1}^{n}\left(-2{W}({x}_{i}-{c})+2r\frac{{I}_{\mathcal{I}}\sqrt{{W}}({x}_{i}-{c})}{\sqrt{({x}_{i}-{c})^{T}\sqrt{{W}}^{T}{I}_{\mathcal{I}}\sqrt{{W}}({x}_{i}-{c})}}\right)={0}.
	\]
	Observe that if $j\in\mathcal{I}$, the $j$-th coordinate of
	the second term is zero, so we have ${c}_{j}=\frac{1}{n}\sum_{i=1}^{n}{x}_{i,j}$
	for any $j\in\mathcal{I}^{c}$. For $j\in\mathcal{I}$, an analytic solution is difficult to find, but gradient descent can provide a numerical solution. 
\end{enumerate}

So far, we have assumed that the underlying support $S_{\mathcal{I}}$
has coordinate axes that are parallel to the coordinate axes. To make
this assumption more realistic, we propose to rotate the dataset $\mathcal{X}$ so
that it can be viewed in a position such that its axes are parallel to
the coordinate axes. Then, we solve the optimization problem \eqref{eq:basic opt with c and r} for the rotated dataset and rotate it back to obtain
reduced dataset. The rotation can be chosen according to the types of datasets as appropriate in the procedure, as is discussed in Sec.\ref{sec: Parameter Tuning}.  

The exhaustive binary search over $2^{d}$ possible subsets is computationally
expensive. We observe that the core of the optimization problem
lies in the subset selection of index set $\{1,\cdots,d\}$. We can rephrase
the optimization problem (\ref{eq:basic opt with c and r}) as follows:
\begin{align}
	\min_{{c}\in\mathbb{R}^{d},r\in\mathbb{R}^{+}}& \sum_{i=1}^{n}\left(({x}_{i}-{c})^{T} {W}({x}_i-{c})+r^{2}\right.\nonumber\\
	& \left. -2r\sqrt{\left({x}_{i}-{c}\right)^{T}\sqrt{{W}}^{T}{v}^{T}{I}{v}\sqrt{{W}}\left({x}_{i}-{c}\right)}\right) & \text{, s.t. }\|{v}\|_{l_{0}}=d'+1,\label{eq:SPCA_standard}
\end{align}
Since $l_{0}$-norm is not convex, solving this problem requires a brute-force step in finding optimal ${v}$ whose entries are either 0 or 1 (and hence $\mathcal{I}$ since ${v}^T{I}{v}={I}_{\mathcal{I}}$), as detailed in Algorithm \ref{alg:SPCA_L1}. 
%However, this does not address the computational issue. 
Instead of using $l_0$ directly in the original problem, we consider the following computationally cheaper alternative:
%{\scriptsize
\begin{align}
	\min_{{c}\in\mathbb{R}^{d},r\in\mathbb{R}^{+}} & \sum_{i=1}^{n}\left(({x}_{i}-{c})^{T} {W}({x}_i-{c})+r^{2}\right.\nonumber\\
	& \left. -2r\sqrt{\left({x}_{i}-{c}\right)^{T}\sqrt{{W}}^{T}{v}^{T}{I}{v}\sqrt{{W}}\left({x}_{i}-{c}\right)}\right) & \text{, s.t. }\|{v}\|_{l_{k}}\leq d'+1,\label{eq:SPCA_L1}
\end{align}
%}
where $k\geq1$ for a convex surrogate norm (usually $l_{1}$ is sufficient). This kind of relaxation is proposed in optimization \citep{boyd2004convex}. As a practical suggestion, when there are more than 500 combinations of binary indices to search exhaustively, we recommend $\ell_1$ relaxation as in \eqref{eq:SPCA_L1}, otherwise we perform an exhaustively search.
For very
high-dimensional dataset, the empirical performances for $l_{0}$
and $l_{1}$ penalties are similar.

\subsection{SRCA Method}\label{sec: SPCA section}
{\singlespacing
\LinesNumberedHidden
\setcounter{algoline}{0}
\begin{adjustbox}{width=1.\textwidth,center}
	\begin{algorithm}[H]
		\caption{\label{alg:SPCA} SRCA dimension reduction algorithm} 
		\KwData{$X$ (data matrix consisting of $n$ samples in $\mathbb{R}^d$)} 
		\KwIn{$d'$ (the dimension of the sphere), ${W}$ (the covariance weight matrix, by default ${W}={I}_d$), $\lambda$ (optional, the sparse penalty parameter), rotationMethod (the method we use to construct the rotation matrix).}
		\KwResult{
			$\hat{{c}}$ (The estimated center of $S_{\mathcal{I}}$ in $\mathbb{R}^d$), $\hat{r}$ (The estimated radius of $S_{\mathcal{I}}$), $\mathcal{I}_{opt}$ (The optimal index subset) 
		}

		%\SetKwFunction{summ}{sum} 
		\SetKwFunction{GetRotation}{GetRotation} \GetRotation($X$,rotationMethod) obtains a $d\times d$ rotation matrix based on the data matrix $X$. The rotationMethod option specifies what method we use to construct the rotation matrix, by default, we use PCA to obtain a rotation matrix. \\
		\SetKwFunction{ProjectToSphere}{ProjectToSphere} \ProjectToSphere($X$,${c}$,$r$,$k$) is a projection that projects the point set $X$ onto an $l_k$-sphere of center ${c}$ and radius $r$ via $X\mapsto{c}+\frac{X-{c}}{\|X-{c}\|_{l_k}}\cdot r$.\\
		
		\Numberline \Begin { 
			\Numberline Standardize the dataset by subtracting its empirical mean $X$ = $X-\bar{X}$ \\
			\Numberline Construct a rotate matrix $R$ = \GetRotation($X$,rotationMethod) \\     
			\Numberline $X_{rotated} = X* R$ \\
			\Numberline $\mathscr{L}_{opt} = \infty$ \\
			\While{$\mathcal{I}\subset{1,\cdots,p}$}{ 
				\Numberline Solve the optimization problem (\ref{eq:SPCA_standard}) with respect to ${c},r$ with a fixed $\mathcal{I}$\\
				\Numberline Denote the solution as ${c}_{cur},r_{cur},\mathcal{I}_{cur}$ \\
				\uIf{$\mathscr{L}({c},r,\mathcal{I}\mid \mathcal{X}) \leq \mathscr{L}_{opt}$}{
					$\mathscr{L}_{opt}\leftarrow \mathscr{L}({c},r,\mathcal{I}\mid\mathcal{X})$ \\
					${c}_{opt}\leftarrow {c}_{cur}$ \\
					$r_{opt}\leftarrow r_{cur}$ \\
					$\mathcal{I}_{opt}\leftarrow \mathcal{I}_{cur}$ \\
				}
			}
			\Numberline Construct the binary index vector ${\eta}=(\eta_i),\eta_i=1\text{ iff }i\in\mathcal{I}$ and $\eta_i=0$ otherwise. \\
			\Numberline $\hat{{c}} = {c}_{opt}\cdot{\eta}*R^{-1}+\bar{X}$ \\
			\Numberline $\hat{r} = r_{opt}$ \\
			\Numberline $X_{rotated}(:,\mathcal{I}) \leftarrow {0}$ \\
			\Numberline $X_{rotated}\leftarrow$ \ProjectToSphere($X$,$\hat{{c}}$,$\hat{r}$,$k$) \\
			\Numberline $X_{output} \leftarrow X_{rotated}*R^{-1} + \bar{X}$ \\
		}
	\end{algorithm}
\end{adjustbox}
}

The method discussed above is refereed to as the \emph{spherical rotation dimension reduction} (SRCA) method and presented in Algorithm \ref{alg:SPCA}, which employs geometric loss functions designed for spherical datasets. The key steps of our proposed SRCA dimension reduction method can be summarized into a ``Rotate-Optimize-Project'' scheme as follows, with $l_1$ algorithms detailed in Supplement \ref{sec:SPCA-Algorithm}. 

%\begin{itemize}
	%\item 
 \textbf{Rotate: Conduct the rotation.} With the chosen rotation method, we construct a rotation matrix ${R}$ based on the dataset $\mathcal{\ensuremath{X}}$. We translate and rotate the dataset $\mathcal{X}$ to a standard position $(\mathcal{X}-\bar{\mathcal{X}}){R}$, so that we can reasonably assume that the axes of the ellipsoid are parallel to the coordinate axes \citep{jolliffe1995rotation}. 
 
\textbf{Optimize: Solve the optimization for the best $d'+1$ axes.} We perform	dimension reduction based on the geometric loss function discussed above. 
As stated in (\ref{eq:SPCA_standard}), we conduct dimension reduction by minimizing the loss function based on the point-to-sphere distance to the estimated sphere $S_{\mathcal{I}}$, to obtain the optimal ${v}_{\text{opt}}$ and the optimal index set $\mathcal{I}_{\text{opt}}$.
	
	\begin{align}\left({v}_{\text{opt}},{c}_{\text{opt}},r_{\text{opt}}\right)= & \arg\min_{{c}\in\mathbb{R}^{d},r\in\mathbb{R}^{+}}\sum_{i=1}^{n}\left(({x}_{i}-{c})^{T} {W}({x}_i-{c})+r^{2}\right.\nonumber\\
		& \left.-2r\sqrt{\left({x}_{i}-{c}\right)^{T}\sqrt{{W}}^{T}{v}^{T}{I}{v}\sqrt{{W}}\left({x}_{i}-{c}\right)}\right)\label{eq:SPCA_standard-1} \text{ s.t. }\|{v}\|_{l_{0}}=d'+1,
	\end{align}
	where the constraint can be relaxed by $\|{v}\|_{l_{k}}\leq d'+1$.
 
	\textbf{Project: project onto the optimal sphere.} 
	Now we project the datapoints back into full space with the chosen dimension and axes, placing  ${x}_{i}$ back onto the sphere $S({c}_{\text{opt}},r_{\text{opt}})$ with the estimated center ${c}_{\text{opt}}$ and radius $r_{\text{opt}}$, the SRCA projection is given by:
	\begin{align}
		\widehat{x}_{i}={c}_{\text{opt}}\cdot{v_{\text{opt}}}+r_{\text{opt}}\frac{({x}_{i}-{c}_{\text{opt}}) {v_{\text{opt}}^T I v_{\text{opt}} }}{\|({x}_{i}-{c}_{\text{opt}}){v_{\text{opt}}^T I v_{\text{opt}}}\|}.
		\label{eq:project-to-sphere}
	\end{align}
%\end{itemize}

% We briefly demonstrate the difference between PCA and SRCA %in Figure \ref{pca-spca_orthogonalLoops} using the example we discussed in Figure \ref{fig:cc-pca_cc}, with more examples featuring realistic noises 
% in Supplement \ref{sec:orthogonalLoop}. It becomes evident that that SPCA and SRCA method better respect the topology of the original dataset under the same $d'$. PCA does not retain the circular structure, but SRCA puts both circles onto a larger 1-sphere congruent to $\mathbb{S}^1$.

\section{Theoretical Results}\label{sec: theoretical results}

We have established the procedure for our propose method, SRCA, in an algorithmic way. Next, we discuss and provide some theoretical results that guarantee the performance of SRCA in applications. Proofs are deferred to supplementary materials, but we want to emphasize that the techniques of $\rho$-loss \citep{huber1967behavior} and $\Gamma$-convergence \citep{braides2002gamma} are introduced to tackle probabilistic properties for DR methods.

\subsection{Convergence}\label{sec: Convergence}
Unlike SPCA, SRCA does not have a closed form solution (i.e., analytic expression of center and radius estimates in terms of dataset $\mathcal{X}$) but relies on the solution to an optimization problem. Therefore, the convergence of this optimization becomes central in our theory development. We briefly discuss the convergence guarantee for the algorithm we designed. In the binary search situation, for each fixed choice of indices, we compute the gradient of loss function. 
With mild assumptions, gradient descent provides linear convergence. 
If the optimization problem (\ref{eq:SPCA_standard-1})
has solutions, then the solution is clearly unique. This is because
there are only finitely many ${v}$ such that $\|{v}\|_{l_{0}}=d'+1$,
and the binary search in the standard algorithm would exhaustively search all possible values
of ${v}$. 

To these ends, we provide a basic convergence for a sub-problem in our Algorithm \ref{alg:SPCA} via the gradient descent algorithm of positive constant step size. The sub-problem is defined by the following loss function:
\begin{align}
	\mathscr{L}_{{v}}({c},r|\mathcal{X},W)=\mathscr{L}_{v}(c,r)\coloneqq & \sum{}_{i=1}^{n}\left(({x}_{i}-{c})^{T} {W}({x}_i-{c})+r^{2} \right. \\
	& \left. -2r\sqrt{\left({x}_{i}-{c}\right)^{T}\sqrt{{W}}^{T}{v}^{T}{I}{v}\sqrt{{W}}\left({x}_{i}-{c}\right)}\right)
\end{align}
The following theorem guarantees the convergence of SRCA.  
\begin{theorem}\label{thm:Lip}
	For a fixed vector ${v}$ (or equivalently ${I}_{\mathcal{I}}$), if we assume that $\|{x}_{i}-{c}\|\leq R_{1},~\forall i=1\cdots,n,~r\leq R_{2}$
	and $|\lambda_{\max}({W})|\leq R_{3}$, where $\lambda_{\max}(\cdot)$ denotes the largest eigenvalue, then for a positive finite constant step size independent of the iteration number $k$, the gradient descent algorithm (c.f., the setting in  \citet{boyd2003subgradient,boyd2004convex}) converges to the optimal value in the following sense, 
	$$\lim_{k\rightarrow \infty}\mathscr{L}_{{v},k}\rightarrow \mathscr{L}_{{v}}^*$$
	where $\mathscr{L}_{{v},k}=\mathscr{L}_{{v}}\left({c}_{k},r_{k} \right)$, $\left({c}_{k},r_{k} \right)$ is the value in the $k$-th iterative step in the gradient descent algorithm, and $\mathscr{L}_{{v}}^*$ denotes the minimum of the loss function for this fixed ${v}$.
\end{theorem}
These results justify that for a fixed ${v}$ (or equivalently $\mathcal{I}$) we can solve the sub-problem defined by the above function, and since we conduct an exhaustive search for the index vector ${v}$, we can find the solution to the original problem (\ref{eq:SPCA_standard-1}) as well.

\subsection{Consistency}
In this section, we assume the observed data are from a ``true'' but unknown sphere $S_{I_0}(c_0,r_0)$ and show that the solution of SRCA is consistent, that it, we can find the true sphere as long as we have enough samples.

\begin{theorem}\label{thm:consistency_clean}
	Assume ${x}_i\in S_{\mathcal{I}_0}(c_0,r_0),~\forall i =1,\cdots,n$ and  $n>d'+1$. Let $\widehat{\mathcal{I}}_k,\widehat{c}_k,\widehat{r}_k$ be the solution of SRCA after $k$ iteration in the solution of the corresponding optimization problem, then
	$$(\widehat{\mathcal{I}}_k,\widehat{c}_k,\widehat{r}_k)\xrightarrow[]{k\to\infty}(\mathcal{I}_0,{c_0},r_0). $$
\end{theorem}

However, the assumption that are observations are exactly on a sphere is unrealistic in practice, as the data often come with measurement errors. Instead, we adopt following (common) assumption in manifold estimation: $x_{i}=y_{i}+\epsilon_{i}$, where the unobserved $y_i$'s are exactly from a sphere $S(c_0,r_0)$ and $\epsilon_i$ represents the measurement error. The next theorem fills in the gap using the $\Gamma$-convergence \citep{braides2002gamma}, which is first applied in DR problems.

\begin{theorem}\label{thm:consistency_noise}
Under the following assumptions:

(A0) The index vector $\mathcal{I}$ is
	fixed and the parameter $\theta=({c},r)\in\Theta\coloneqq[-C,C]^{d}\times[R_0,R]\subset\mathbb{R}^{d}\times\mathbb{R}^{+}$
	for some $C,R_0,R>0$.  

(A1) $x_i$'s are compactly supported. 

(A2) $\lim_{n\to\infty}\frac{1}{n}\sum_{i=1}^{n}\|\epsilon_{i}\|=0$

the SRCA solution $\widehat{\theta}_n\to \theta_0$ as $n\to\infty$. 

\end{theorem}
In other words, the SRCA estimator based on noisy samples is consistent, that is, converges to the true parameter $\theta_0$, as low as the noise decays to zero with sample size. In fact, this assumption is even weaker than those in existing literature, see \cite{maggioni2016multiscale,fefferman2018fitting,aamari2019nonasymptotic} for more details. For example, in \cite{aamari2019nonasymptotic} the amplitude of the noise is assume to be $\|\epsilon\|\sim n^{-\frac{\alpha}{d}}$ for $\alpha>1$. In contrast, we only require $\|\epsilon\|\to0$, so $\|\epsilon\|\sim n^{-\alpha}$ for any $\alpha>0$ or even $\|\epsilon\|\sim \frac{1}{\log n}$ is good enough.

\subsection{Asymptotics}\label{sec: Asymptotics of SPCA}
In this section, we consider the asymptotic behavior of SRCA optimization result when the underlying dataset is assumed to be drawn from a probabilistic distribution, regardless whether it is supported on a sphere or not.

To yield the asymptotic results, we need to take the perspective of robust statistics as mentioned in the end of Section \ref{sec: sparse penalty}.  The asymptotic theory here is a specific case of empirical risk minimization. With a mild technical assumption that the parameter $\theta=({c},r)\in\Theta\coloneqq[-C,C]^{d}\times[R_0,R]\subset\mathbb{R}^{d}\times\mathbb{R}^{+}$
for some $C,R_0,R\in(0,\infty)$, our loss function and optimization problem can be expressed as 
\begin{align}
	& \min_{\mathcal{I}\subset\{1,\cdots,d\},{c}\in[-C,C]^{d}\subset\mathbb{R}^{d},r\in[R_0,R]\subset\mathbb{R}^{+}}\mathscr{L}({c},r,\mathcal{I}\mid\mathcal{X},{W})\label{eq:3.4}\\
	& =\min_{\mathcal{I}\subset\{1,\cdots,d\},{c}\in[-C,C]^{d}\subset\mathbb{R}^{d},r\in[R_0,R]\subset\mathbb{R}^{+}}\sum_{i=1}^{n}\left(({x}_{i}-{c})^{T} {W}({x}_i-{c})+r^{2}\right.\label{eq:robustEQ}\\
	& \left. -2r\sqrt{\left({x}_{i}-{c}\right)^{T}\sqrt{{W}}^{T}{I}_{\mathcal{I}}\sqrt{{W}}\left({x}_{i}-{c}\right)}\right) \text{s.t. }|\mathcal{I}|=d'+1\nonumber
\end{align}

For a fixed $\mathcal{I}$, \eqref{eq:robustEQ} can be written in the form of (3.1) in \citet{huber2004robust}, i.e., 
\[
\rho({x};\theta)=\left(({x}_{i}-{c})^{T} {W}({x}_i-{c})+r^{2}-2r\sqrt{\left({x}-{c}\right)^{T}\sqrt{{W}}^{T}{I}_{\mathcal{I}}\sqrt{{W}}\left({x}-{c}\right)}\right)
\]
Correspondingly, we can write Huber's $\psi$-type function of $\rho$ as $\psi(\theta)=\frac{\partial\rho(\theta)}{\partial\theta}$.

Classical style asymptotic results are presented below in Theorem \ref{Theorem A}, which  states that, with mild assumptions, the estimates $T_n$ obtained by solving SRCA would estimate the center and radius of the spherical space consistently, corresponding to Huber's $\rho$-type estimator consistency \citep{huber1967behavior}; Theorem \ref{Theorem B} states that with more stringent conditions on continuity of $T_n$, asymptotic normality of these estimators can also be formulated into Huber's $\psi$-type normality.

To apply these two asymptotic results, we need to make mild assumptions on the parameter space and assume that we already know the retained dimension $d'+1$. 
\begin{theorem}\label{Theorem A}Suppose (A0) in Theorem \ref{thm:consistency_noise} holds and the samples ${x}_{1},\cdots,{x}_{n}\in\mathbb{X}=\mathbb{R}^{d}$
	of size $n$ are i.i.d. drawn from the common distribution $P$. $P$ has
	finite second moments on the probability space $(\mathbb{X},\mathcal{A},\nu)$
	with Borel algebra $\mathcal{A}$ and Lebesgue measure $\nu$. Then
	the consistent estimator $T_{n}$ for parameter $\theta=({c},r)$ defined by 
	\begin{align*}
		\frac{1}{n}\sum_{i=1}^{n}\rho({x}_{i};T_{n})-\inf_{\theta\in\Theta}\frac{1}{n}\sum_{i=1}^{n}\rho({x}_{i};\theta) & \xrightarrow{n \rightarrow\infty}0,\text{a.s. }P%\\
		%n & \rightarrow\infty,
	\end{align*}
	would converge in probability and almost surely to $\theta_{0}$ w.r.t.
	$P$ (for the true parameter values $\theta_{0}$ defined on page \pageref{true parameter}). Particularly, $T_{n}$ can be realized as a solution to our
	optimization problem (\ref{eq:3.4}) above. %$\square$
\end{theorem}
Unlike Theorem \ref{thm:Lip}, which concerns the convergence of the algorithm, we assume that the fixed i.i.d. samples $\mathcal{X}$ are drawn from a probability distribution. Similarly, we have a distributional result as follows. 
%\textbf{Theorem B.} 
\begin{theorem}\label{Theorem B}
	In addition to the assumption in Theorem \ref{Theorem A}, we assume $P\left(\left|T_{n}-\theta_{0}\right|\leq\eta\right)\rightarrow1$
	as $n\rightarrow\infty$, then the estimator $T_{n}$ defined by 
	\begin{align*}
		\frac{1}{\sqrt{n}}\sum_{i=1}^{n}\psi({x}_{i};T_{n}) & \xrightarrow{n\rightarrow\infty}0,\text{a.s. }P%\\
		%n & \rightarrow\infty,
	\end{align*}
	would satisfy 
	%\begin{align*}
 $
		\frac{1}{\sqrt{n}}\sum_{i=1}^{n}\psi({x}_{i};T_{n}) %& 
  +\sqrt{n}\lambda(T_{n})\xrightarrow{n \rightarrow\infty}0,\text{a.s. }P%\\
		%n & \rightarrow\infty,
	%\end{align*}
 $
	Particularly, our loss function would satisfy differentiability at
	$\theta_{0}$ and $\sqrt{n}\left(T_{n}-\theta_{0}\right)$ is asymptotically
	normal with mean zero and covariance matrix $$\left(\nabla_{\theta_{0}}\lambda^{-1}\right)\cdot\left(\left[\psi({x}_{i};\theta_{0})-\mathbb{E}_{P}\psi({x}_{i};\theta_{0})\right]^{T}\left[\psi({x}_{i};\theta_{0})-\mathbb{E}_{P}\psi({x}_{i};\theta_{0})\right]\right)\cdot\left(\nabla_{\theta_{0}}\lambda^{-1}\right)^{T}.$$
	%$\square$
\end{theorem}
Defined by the geometric loss function $\mathscr{L}$, SRCA does not have an analytic solution, but this loss benefits from the theoretic results above and can be replaced by other types of loss functions, enabling SRCA to be applied more widely.

Note that we also assumed that the index set $\mathcal{I}$ is fixed for our statements of theorems. In the exhaustive search, these results above can be applied individually to fixed $\mathcal{I}$; but in the $l_1$-relaxed problem \eqref{eq:SPCA_standard-1}, since the optimization is a joint optimization our  asymptotic results Theorem \ref{Theorem A} and \ref{Theorem B} in this section do not apply. 

\subsection{Loss Function Minimization}\label{sec: MSE comparison}
Next, we consider the theoretic behavior of SRCA in terms of approximating a general manifold. The following theorem compares the MSE of PCA, SRCA (when the rotation is chosen by PCA) and SPCA:

\begin{theorem}\label{thm: MSE comparison}Given data ${x}_1,\cdots,{x}_n$ in a bounded subset of $\RR^d$, let $H$ be the best subspace obtained by PCA, $S_1$ be the sphere obtained by SPCA and $S_2$ be the sphere obtained by SRCA with rotation provided by PCA, then
	$$\sum_{i=1}^n d^2({x}_i,S_2)\leq \min\left\{\sum_{i=1}^n d^2({x}_i,H),\sum_{i=1}^n d^2({x}_i,S_1)\right\}.$$
\end{theorem}

That is, SRCA has the best approximation performance in terms of MSE among PCA, SPCA and SRCA, regardless of the true support of the observations.

To summarize and interpret our theoretical results briefly, Theorem \ref{thm:Lip} ensures that a gradient-descent algorithm can be used for solving the loss function minimization problem \eqref{eq:basic opt with c and r} for any finite samples with convergence guarantees; Theorem \ref{thm:consistency_clean} and \ref{thm:consistency_noise} show that SRCA can recover the true sphere, if it exists, when the data are clean or with measurement error. For general case where the observations are not necessarily supported by a sphere, Theorems \ref{Theorem A} and \ref{Theorem B} ensure that the sequence of finite-sample minimizers of our loss function asymptotically converges to minimizer $\theta_0$; Theorem \ref{thm: MSE comparison} points out that SRCA can better approximate the unknown support in terms of MSE than PCA and SPCA. 

\section{Numerical Experiments}\label{sec: Numerical Experiments}
With theoretical results above on MSE, we also wish to examine the practical performance of SRCA against the state-of-the-art dimension reduction methods on real datasets. We focus on the empirical structure-preserving and coranking measurements below, an application to the motivating dataset about cell cycle, and discuss the choice of parameters in SRCA in the end. Details of selected datasets are in Supplement \ref{sec: Dataset Selection}.

\subsection{MSE\label{sec:MatchedDistMSE}}

As a dimension reduction method, the most common and natural measurement of performance is based on the mean squared error (MSE) between the original and reduced datasets, which measures how close the manifold is to the original observations. However, most dimension reduction methods only output low dimension features, like LLE, Isomap, tSNE, UMAP, GPLVM, etc, where the MSE is not well-defined because the low dimensional features cannot be trivially embedded into original data space $\mathbb{R}^d$. Algorithms that output projected data in the original space $\RR^d$ include SPCA, PCA, and our proposed SRCA.

Table \ref{table:MSE_BIG_TABLE} presents the MSEs of three competing algorithms on these datasets with $d'=\min\{d-1,4\}$. The out-sample MSEs show a similar patterns and  is postponed to the Supplement \ref{sec:out-of-sample MSE}. It is evident that SRCA has the property of MSE minimization for most datasets and most $d'$, as predicted by the theory in Theorem \ref{thm: MSE comparison}.

\begin{table}[ht!]
	\centering 
	\global\long\def\~{\hphantom{0}}%
	
	%\begin{turn}{90}
	\begin{tabular}{cccccc}
		Dataset
		& Method/$d'=$ & 1 & 2 & 3 & 4\tabularnewline
		\midrule
		\midrule 
		\multirow{3}{*}{Banknote} & PCA & 15.6261 & 6.3356 & 1.9479 & \tabularnewline
		\cmidrule{2-6} 
		& SPCA & 16.3717 & 8.1004 & 1.7348 &    \tabularnewline
		\cmidrule{2-6}  
		& SRCA & \bf{13.439} & \bf{5.5088} & \bf{1.0743} &    \tabularnewline
		\midrule
		Power & PCA & 222.2971 & 55.4460  & 23.5173 & \bf{2.9957}  \tabularnewline
		\cmidrule{2-6}  
		Plant & SPCA & 162.8865  & 102.1006  & 45.5793  & 41.5251     \tabularnewline
		\cmidrule{2-6} 
		& SRCA & \bf{150.8041}  & \bf{52.1439}  & \bf{19.8839}  & 3.0373  \tabularnewline
		\midrule 
		\multicolumn{1}{c}{User } & PCA & 0.1921 & 0.1253  & 0.0718  & 0.0311   \tabularnewline
		\cmidrule{2-6}  
		Knowledge & SPCA & 0.1465  & 0.0893  & 0.0477  & 0.0148  \tabularnewline
		\cmidrule{2-6}  
		& SRCA & \bf{0.1458}  & \bf{0.0887}  & \bf{0.0471}  & \bf{0.0142}   \tabularnewline
		\midrule 
		\multirow{3}{*}{Ecoli} & PCA & 0.076693  & 0.035222  & 0.020522  & \bf{0.00756}   \tabularnewline
		\cmidrule{2-6} 
		& SPCA & \bf{0.047776}  & 0.032948  & 0.019648  & 0.01136    \tabularnewline
		\cmidrule{2-6}  
		& SRCA & 0.076660  & \bf{0.032799}  & \bf{0.018332}  & \bf{0.00756}   \tabularnewline
		\midrule
		\multirow{3}{*}{Concrete} & PCA & 6.7056  &  4.7711  &  3.3636 &  2.3218 \tabularnewline
		\cmidrule{2-6}  
		& SPCA & 5.4743   & 3.9934  &  2.9681   & 1.9181  \tabularnewline
		\cmidrule{2-6} 
		& SRCA & \bf{5.4741}  &  \bf{3.9909}  &  \bf{2.9552}   & \bf{1.8985}   \tabularnewline
		\midrule 
		\multirow{3}{*}{Leaf} & PCA & 0.0245  &  0.0121    &0.0059&    0.0038  \tabularnewline
		\cmidrule{2-6} 
		& SPCA & \bf{0.0155}  &  0.0094  &  0.0072  &  0.0037 \tabularnewline
		\cmidrule{2-6} 
		& SRCA & 0.0230  &  \bf{0.0093}  &  \bf{0.0054}  &  \bf{0.0033}
		\tabularnewline
		\midrule 
		\multirow{3}{*}{Climate} & PCA & 1.4100  &  1.3204  &  1.2323  &  1.1450  \tabularnewline
		\cmidrule{2-6} 
		& SPCA & 1.3563  &  1.2648 &   1.1781  &  1.0907 \tabularnewline
		\cmidrule{2-6} 
		& SRCA & \bf{1.3554} &   \bf{1.2646}  &  \bf{1.1780}  &  \bf{1.0905}
		\tabularnewline
		\bottomrule
	\end{tabular}
	\caption{MSE for different experiments.}
	\label{table:MSE_BIG_TABLE}
\end{table}

\subsection{Cluster Preserving }\label{sec: Structure Preserving}
Cluster structure properties of different dimension reduction algorithms varies, 
however, we hope that the data points  belonging to the same group in the original dataset, are close together in the dimension-reduced dataset. 

For visualization purposes, we fix retained dimension to be $d'=2$ and compare the following six algorithms: SRCA, SPCA, PCA, LLE, tSNE, UMAP. 
We choose these state-of-the-art competitors to visualize in 2-dimensional figures. To further quantify the how well the clustering structures are preserved, the Silhouette Score (SC, \citep{rousseeuw1987silhouettes}), Calinski-Harabasz Index (CHI, \citep{calinski1974dendrite}) and Davis-Bouldin index (DBI, \citep{davies1979cluster}) are considered. Higher SC and CHI, lower DBI imply better separation between clusters in the dataset. We provide these measures on the original labeled dataset (without any DR) as baselines.

\begin{figure}
	\centering
	\includegraphics[width=1\textwidth,height = 200pt]{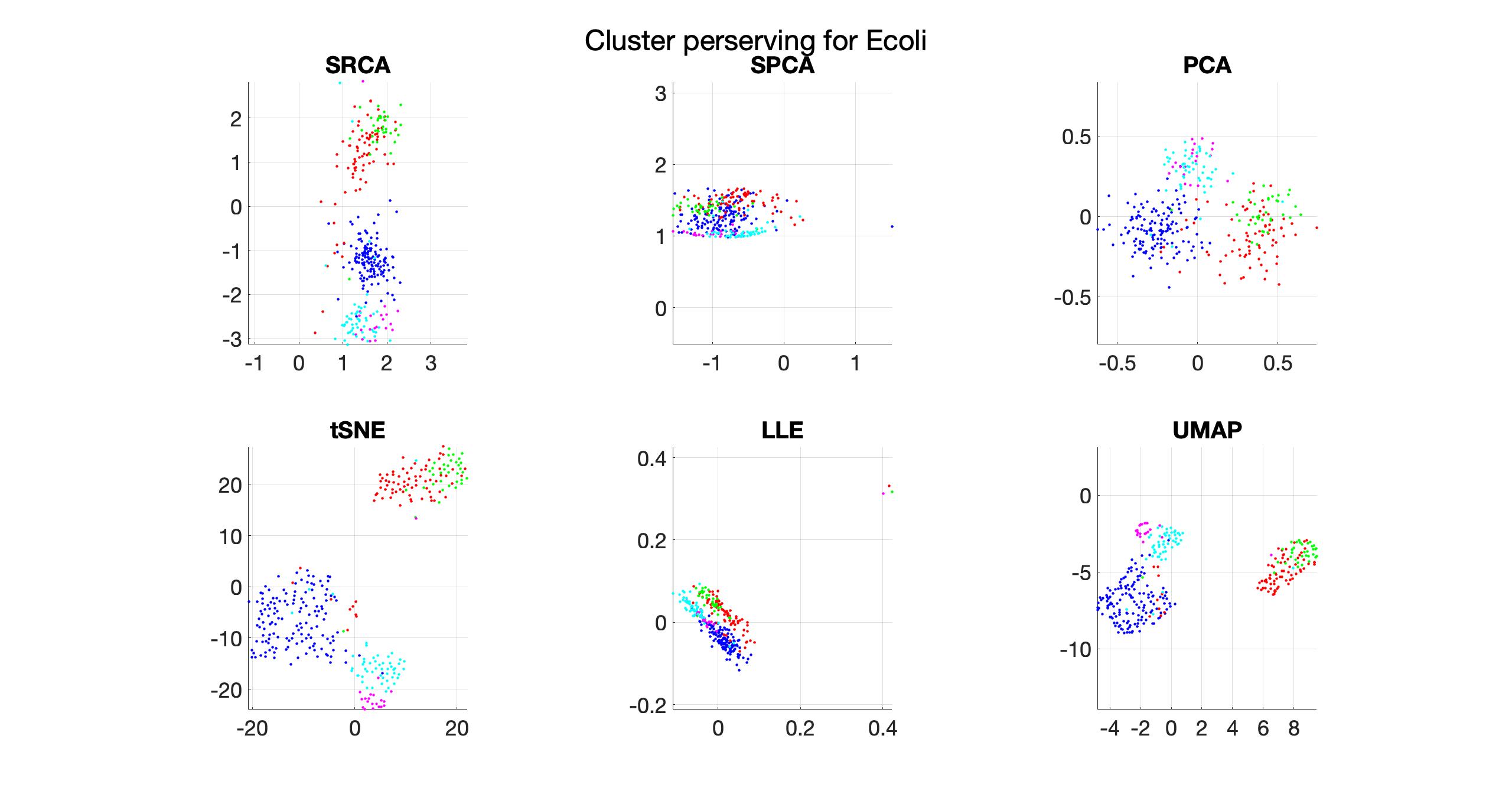}
	\caption{Cluster structures for Ecoli, $d'=2$, there are five different clusters represented by different colors in the reduced dataset. }
	\label{fig:Ecoli_cluster}
\end{figure}

Figure \ref{fig:Ecoli_cluster} and Table \ref{table:Ecoli_cluster} show that SRCA outperforms SPCA, PCA and LLE in terms of all three metrics and comparable to tSNE and UMAP. SRCA also has advantage over its predecessor SPCA and simpler linear method like PCA. With these experiments and Supplement \ref{sec:basic-example}, we conclude that if the dataset has strong spatial sphericity, we usually have good cluster preserving properties from SRCA. If the dataset is highly non-linear, tSNE and UMAP are usually better at the cost of 
creating fake clusters if tuning parameters are not well-chosen~\citep{wattenberg2016how,wilkinson_distance-preserving_2020}.

\begin{table}[h!]
	\centering 
	\global\long\def\~{\hphantom{0}}%
	
	\begin{tabular}{ccccccccc}
		\hline 
		Index & Baseline & SRCA & SPCA  & PCA & LLE &tSNE &UMAP \tabularnewline
		\hline
		SC &0.257& 0.267& 0.260& 0.200& 0.209& 0.293&0.290\tabularnewline
		\hline
		CHI & 133 & 192 & 190 & 215 & 46.6& 376 & 376\tabularnewline
		\hline 
		DBI & 1.49& 1.59& 1.58& 2.56& 2.40&  1.37&1.32\tabularnewline
		\hline 
		
	\end{tabular} 
	\caption{Clustering performance measures for Ecoli}\label{table:Ecoli_cluster} 
\end{table}

\subsection{Coranking Matrix}\label{sec: Coranking matrix}
Another type of quantitative measures is based on the \emph{coranking matrix}
\citep{lee2009quality,lueks2011evaluate}. The coranking matrix can be viewed as the joint histogram of the ranks of original samples and the dimension-reduced samples. The coranking matrix can be used to assess
results of dimension reduction methods. Entry $q_{kl}$ in the coranking matrix is defined as $q_{kl}\coloneqq\{(i,j)\mid \rho_{ij}=k\text{ and }r_{ij}=l\},$ where $\rho_{ij}\coloneqq\{k:d({x}_i,{x}_k)<d({x}_i,{x}_j)\text{ or }d({x}_i,{x}_k)=d({x}_i,{x}_j),k<j \}$ stores the rank of the pair ${x}_i,{x}_j$ in the original dataset; $r_{ij}\coloneqq\{k:d(\hat{{x}}_i,\hat{{x}}_k)<d(\hat{{x}}_i,\hat{{x}}_j)\text{ or }d(\hat{{x}}_i,\hat{{x}}_k)=d(\hat{{x}}_i,\hat{{x}}_j),k<j \}$ stores the rank of the pair $\hat{{x}}_i,\hat{{x}}_j$ in the dimension-reduced dataset, where the rank pair reversed in the dimension-reduced dataset. 
An ideal dimension reduction method should preserve all the ranks of these pairwise distances between original and reduced datasets. That is, we have identical ordering of these pairwise distances in the original space and the dimension-reduced space. 
Coranking matrix is a finer summary but is  related to $ijk$ rank test (See, e.g., \citet{solomon2021geometry}).

We provide three scores (the higher the better) related to coranking matrices of the dimension-reduced results: CC (cophenetic correlation, measuring correlation between distance matrices), AUC (area under curve for the $R_{NX}$ score), WAUC (weighted AUC) computed from $\mathtt{coRanking}$ R-package \citep{kraemer2018dimred}.  

\begin{figure}[h!]
	\centering
	\includegraphics[width=0.9\textwidth]{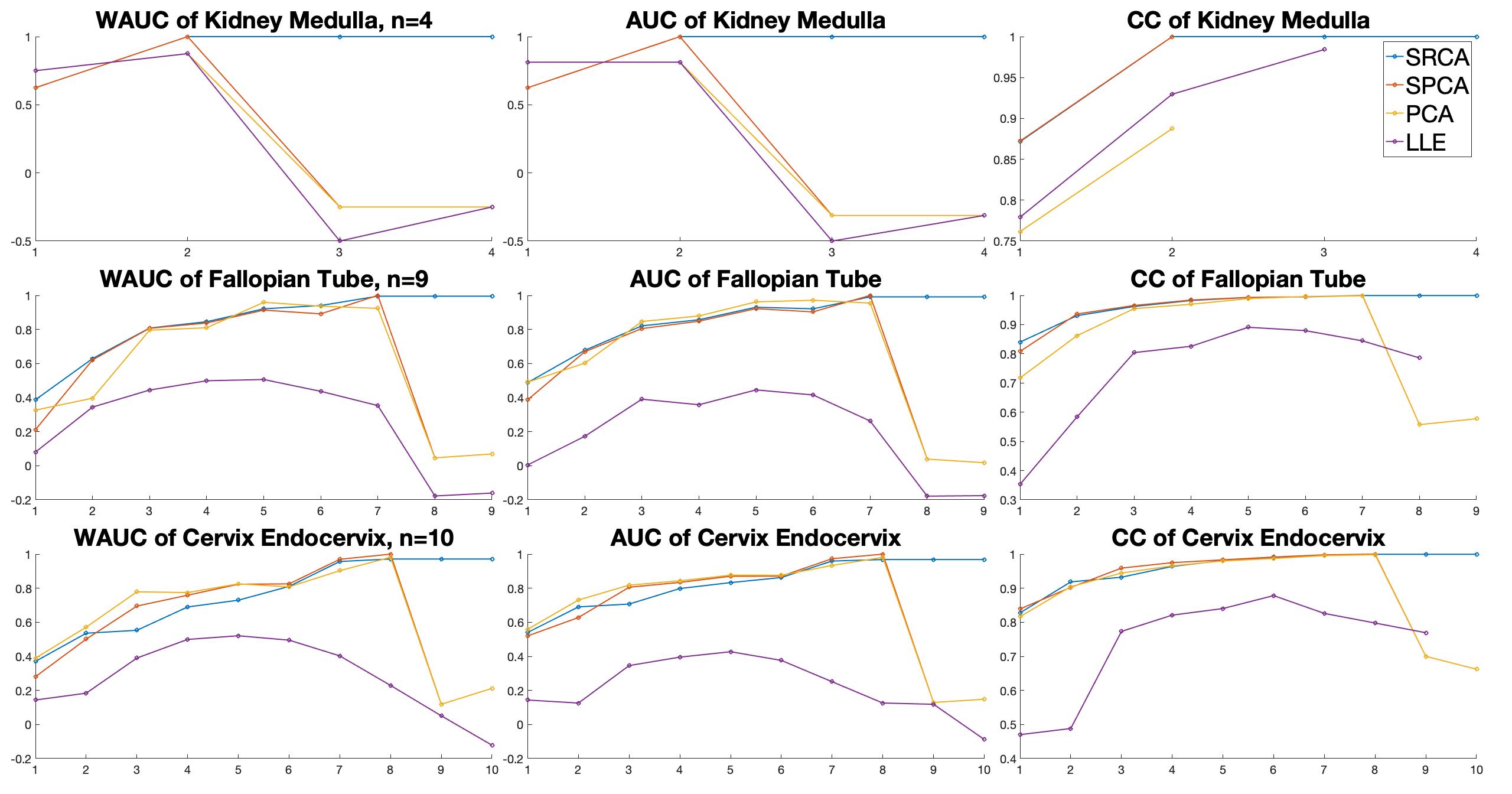}

	\caption{Coranking measurements of three GTEx tissues for different retained dimension, the horizontal axes are retained dimension $d'$, the vertical axes are score values.}
	\label{fig:corankings}
\end{figure}

To understand our subsequent analyses better, we referred our readers to the analysis of the dimension reduction result of simple examples like $\mathbb{S}^2$, $\mathbb{T}^2$ and a plane diffeomorphic to $\mathbb{R}^2$, evaluated by these coranking matrix related scores in Supplement \ref{sec:basic-example}, where SRCA is the only dimension method that consistently behaves almost the best in plane, spheres and topologically non-trivial examples like torus when measured by coranking scores. 
Another advantage of SRCA over existing DR methods is that it allows $n<d'$, which happens to a variety of real datasets, specially for biomedical data where both $d$ and $d'$ are large. For example, in Genotype-Tissue Expression(GTEx) dataset~\citep{gtex2020gtex}, some tissues are hard to collect so the sample sizes are small but the dimension is very high, like Kidney Medulla ($n=4)$, Fallopian Tube ($n=9$) and Cervix Endocervix ($n=10$). However, there are thousands of genes so we expect that the intrinsic dimension $d'>n$. Following the common practice of feature selection in this database, we subsetted the data to the most variable 500 genes~\citep{townes2019feature}. Most competitors mentioned before are not directly applicable anymore when $d'>n$, including tSNE, UMAP, Isomap, MDS, etc. For illustration purpose, we retain the first $n$ dimensions. As a result, we present the three coranking based measurements on three tissues obtained from SRCA, SPCA, PCA and LLE for different $d'$ in  Figure \ref{fig:corankings}.

%From plots in Figure \ref{fig:corankings}, we conclude that SRCA almost always outperforms SPCA, PCA and LLE in terms of coranking measures when $d'<n$. When $d'>n$, none of the competitors can be generalized easily to show performance comparable to SRCA, as expected. 

\subsection{Application:  Human Cell Cycle}\label{sec:cell_cycle}
The human cell cycle consists of four growth phases: G1, S, G2, and M. In a recent study, non-transformed human retinal pigmented epithelial (RPE) cells were genetically engineered to express a fluorescent cell cycle reporter that enables accurate identification of each cell's phase (i.e., G1, S, G2, or M) through time-lapse imaging \citep{stallaert2022structure}. Subsequently, the cells were fixed and subjected to iterative indirect immunofluorescence imaging (4i) to measure 48 key cell cycle effectors in 8,850 individual cells. A total of 246 single-cell features were derived from this imaging dataset, including protein expression and localization (e.g., nucleus, cytosol, perinuclear region, and plasma membrane), cell morphological attributes (such as nucleus and cell size and shape), and microenvironment characteristics (like local cell density), ultimately generating a comprehensive cell cycle signature for each cell within the population.

In their study, \cite{stallaert2022structure} narrowed the features to a set of 40 that most accurately predicted cell cycle phase (refer to Figure S1 panel A in \cite{stallaert2022structure}). Thus, the reduced dataset has a sample size of $n=8,850$ and an ambient dimension of $d=40$. Our goal is to decrease the dimension to $d'=2$ for visualization purposes, while maintaining the four clusters that correspond to the four phases (G1, S, G2, M) and the cyclic structure: G1 → S → G2 → M → G1. To account for the diverse units of the 40 selected features, we applied z-score normalization to the data.

\begin{figure}
	\centering
	\includegraphics[width=1\textwidth,height = 200pt]{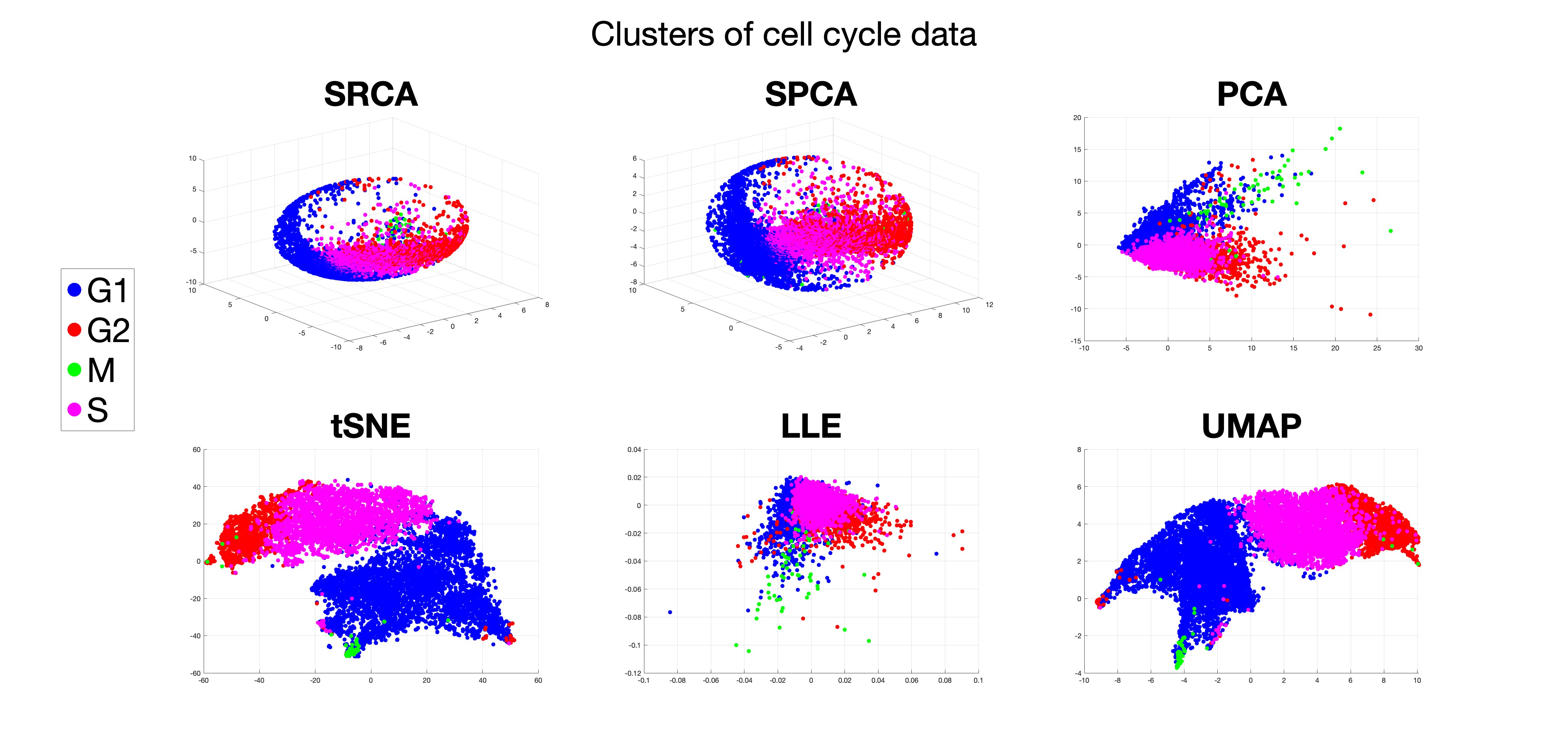}
	\caption{Cluster structures for human cell cycle, $d'=2$, colored by phase. }
	\label{fig:cell_cycle_cluster}
\end{figure}

Figure \ref{fig:cell_cycle_cluster} displays the visualization of cell cycle data, with colors representing different cell cycle phases. While all algorithms can distinguish the four phases, PCA, tSNE, LLE, and UMAP fail to capture the cyclical structure. For example, the green points (M) should be located between the blue points (G1) and red points (G2); and magenta points (S) should be opposite to green points. In contrast, both SRCA and SPCA successfully recover the cyclical structure on 2-dimensional spheres. To compare SRCA and SPCA, we assess the MSE, as shown in the first column of Table \ref{table:cell_cycle}.

\begin{table}[h!]
	\centering 
	\global\long\def\~{\hphantom{0}}%

	\begin{tabular}{cccccc}
		\hline 
		&  MSE & Var(G1)  &Var(S) & Var(G2)&Var(M) \tabularnewline
		\hline 

  PCA & 22.934& 575 & 81& 1297 & 3631\tabularnewline\hline 
		SPCA  & 22.252 &566 & 110& 276 &763\tabularnewline
		\hline 
		SRCA &\bf{22.098} &\bf{633} &117& \bf{434}&\bf{475}\tabularnewline
		\hline
	\end{tabular} \caption{Quantitative metrics of SRCA and SPCA for cell cycle data}\label{table:cell_cycle} 
\end{table}

Given that the true structure is cyclical, clustering metrics that depend on linear structures, such as the Silhouette score, are not suitable for this example. Instead, we use external biological information to validate our findings. Among the four phases, G1 cells are known to possess greater degrees of freedom \citep{chao2019evidence,stallaert2022molecular}, leading us to anticipate that the variance of samples within G1 will be the largest among the four phases, with variances for G2 and M being similar and variance for S being the smallest. The variance is defined as the distance between each sample and the cluster mean, so a larger variance indicates a more dispersed distribution of points within that cluster. The final four columns in Table \ref{table:cell_cycle} demonstrate that SRCA more accurately captures the heterogeneity of cell activities across different phases.

\subsection{Parameter Selection}\label{sec: Parameter Tuning}
It is a separate but important problem to select (or tune) the parameter of both classical and modern DR methods.
For classical DR methods, like PCA or MDS, the parameter usually has a explicit geometric interpretation. For modern non-linear DR methods, like tSNE and UMAP, the parameters affect both reproducibility and interpretability of the resulting dimension-reduced dataset. 

% \subsubsection{Retained dimension \texorpdfstring{$d'$}{d'}}
The first parameter that dictates the behavior of most DR methods is the retained dimension $d'$, which can be determined by subsequent purpose (e.g., the tSNE and UMAP usually take $d'=2,3$ for visualizations). 
%When the practitioner is unsure about the choice of $d'$, there is a trade-off between retained information and reduction of dimensionality. 
% For linear  DR methods like PCA, we usually use the residual norms (i.e., the total variance minus explained variance) to decide $d'$. 
% Multiple cautionary notes \citep{jolliffe1982note} have pointed out that such a straightforward selection of principal components based purely on the magnitude of eigenvalues could be problematic. 

% In our method SRCA, the $d'$ is selected based on MSE computed from a non-linear distance function. Besides its non-linear nature, we tie the criterion back to the geometric nature of the dataset. Therefore, we expect this criteria   naturally preserve distances and  clustering induced by geometry. We also note that the MSE-based criteria is a benefit of approximating the manifold in the original space (SRCA, SPCA, PCA), instead of a new space (LLE, tSNE, etc). 
%\subsubsection{Rotation methods}\label{sec: type of rotation methods}
The second parameter is the choice of rotation methods, which is highly data dependent and affects the clustering and visualization most. Regarding the MSE performance, Table \ref{table:Banknote_MSE_d} investigates the performance of SRCA with different kinds of rotations. We can see that the PCA rotation usually gives a reasonable result in terms of MSE performance. Both PCA and quartimax rotations used along with SRCA method  outperforms dimension reduction of PCA and SPCA separately. We have similar observations for some other datasets with $n>d$ (e.g., Leaf, PowerPlant, etc., see~Supplement~\ref{sec: Dataset Selection}).

The choice of rotation methods could also be made to accommodate the type of noise in observations. In the situation where the tail behavior of the noise is close to Gaussian and the ${W}$ is known (or, by default ${I}$), PCA is our default choice; but in the situation where the noise is non-Gaussian and we do not have much knowledge for ${W}$, then ICA \citep{hyvarinen2000independent} is a better alternative. 

Based on the empirical evidence obtained from real datasets (e.g., Table \ref{table:Banknote_MSE_d}), we recommend using PCA rotation as a default, but other types of rotations can be useful for specific datasets, if desired \citep{jolliffe1995rotation}. 
\begin{table}[h!]
	\centering 
	\global\long\def\~{\hphantom{0}}%
	\smaller{
		\begin{tabular}{ccc|ccccccc}
			\hline 
			& PCA & SPCA  & &   &   &SRCA &  &   \tabularnewline
			\hline 
			$d'$ &  & &  PCA & varimax & orthomax &quartimax & equamax & parsimax & ICA  \tabularnewline
			\hline 
			1  &15.6 & 16.4     & 13.4        &18.4&    18.5  &14.9& 27.5 &26.6 & 14.5\tabularnewline
			\hline 
			2 & 6.34&   8.10  &5.51 & 6.19 &6.19 & 5.79& 13.2 &13.4 & 5.36\tabularnewline
			\hline 
			3  & 1.95&    1.73 &1.07 &1.07& 1.07& 1.07 &1.07& 1.07 & 1.14\tabularnewline
			\hline 
		\end{tabular} \caption{%\hrluonote{Put a ICA column here}. 
			MSE of Banknote (see~Supplement\ref{sec: Dataset Selection}) for different rotation methods in the SRCA procedure. We also include two other DR methods PCA and SPCA to compare against SRCA. The first row records the DR methods (PCA, SPCA and SRCA); the second row records the optimal rotation method used by SRCA.}\label{table:Banknote_MSE_d} 
	}
\end{table}

%\subsection{Observations}\label{sec: Observations of experiments}
%\textbf{Summary.} 
We summarize the observations from above experiments in sections \ref{sec:MatchedDistMSE} to \ref{sec: Parameter Tuning}.
Although SRCA is the slowest in terms of computational time among SRCA, SPCA and PCA, it is rather fast 	compared to some non--linear methods like Isomap.

When the retained dimensions $d'<\min\{d,n\}$, SRCA behaves very similarly to SPCA in terms of the MSE, both outperform PCA alone across different real datasets. The interesting observation is that SRCA out-competes both of them in some simple but geometrically nontrivial examples like the ones in Supplement \ref{sec:basic-example}, especially in clustering tasks. The choice of rotation methods can improve the SPCA but PCA rotation is usually good enough.

When $d>d'\geq n$, SRCA outperforms PCA, SPCA and other non-linear DR methods in terms of the coranking scores across different real datasets. Only SRCA yields consistently better dimension reduction results when $d'<n$ and $d'\geq n$.
\section{Contribution}
In this paper, we propose a novel DR method by proposing a rotation-based method with a geometric-induced loss function that minimizes the point-to-sphere distance from original to target spaces. Our motivation is to get the dimension reduction for spherical datasets (or datasets with spherical and elliptical structures) to respect the geometry in the original space. Its variant also works with a general weight matrix ${W}$ and a sparsity penalty $\lambda$. The proposed method is statistically principled, and is theoretically guaranteed to perform well asymptotically. 

Unlike traditional DR methods like PCA and MDS, SRCA works smoothly with a stable performance even when $d\geq d'+1>n$ which is extremely important in biomedical data DR, especially in gene expression data (e.g.,GTEx). Accompanying generalized algorithms for SRCA are also developed, with detailed convergence and a straightforward parallel potentiality for real-world practice. SRCA is related to PCA and SPCA but also generalizes the former into a spherical setting and the latter one into a one-step procedure. Most importantly, SRCA removes the $d'<n$ requirements in these predecessors in a unified framework using novel loss functions.

Compared to non-linear methods, SRCA has geometrical interpretation and practical convenience. Its unique binary search also allows parallelizations when applied to big datasets. A comprehensive experimental study of SRCA against a collection of state-of-art DR methods has been done with detailed qualitative and quantitative measures, revealing the superiority of SRCA.

%\subsection{Discussions and Future Works}
Our current technique focuses on but is not limited to spherical datasets. Similar designs of loss functions can be generalized to a wider variety of spaces like symmetric spaces \citep{li2020density} using multivariate decomposition technique like ICA, and a wider range of datasets like binary datasets \citep{landgraf2020dimensionality}.
The geometric loss function has some relation to or is motivated by the statistical problem we are trying to solve. Then the reduced dataset should be more suitable for the subsequent statistical method to be applied. 

To sum up, we provide a principled way that targets at an explicitly designed geometric loss function and provide an algorithm along with statistical guarantees and a comprehensive evaluation of this novel DR method.

%In terms of DR research, we point out and give a partial solution to a problem in the research by various novel examples in the text (i.e., Appendices \ref{sec:orthogonalLoop} and \ref{sec:basic-example}): how should we respect the geometry and/or topology of the underlying space when performing DR or subsequent statistical problems and tasks?

There are several directions of future work that we wish to pursue.
%\begin{enumerate}
%\item 
%Visualization using SRCA. We expect to use SRCA for visualization and exploratory data analysis with consistency results such as results in the sense of  \citet{johnstone2009consistency}. Besides the theoretic study of sparsity in high-dimensional dataset, parallel computations for the binary search in high-dimensional datasets is of practical interest. Especially when both $n$ and $d$ are high and when $d'+1$ exceeds $n$, PCA and SPCA fail but SRCA still works with optimal rotations.
For example, it is of great interest to see how geometric or topological loss function DR methods perform in data visualization \citep{sigmund200199,nigmetov2022topological}.
 %Efficient randomized implementation with theoretical guarantees. Note that the novel formulation (\ref{eq:general opt}) of our DR method falls into a special case of ``weighted low-rank'' approximation problems \citep{srebro2003weighted,razenshteyn2016weighted}, which leads us to a possible future direction of leveraging the power of numerical linear algebraic algorithms in SRCA. We expect that this would allow us to perform DR methods on much larger datasets. 
%\end{enumerate}

\subsection*{Acknowledgement}
HL wants to thank Dmitriy Morozov, Leland Wilkinson for motivating discussions and comments on early manuscripts; Yin-Ting Liao for discussions in technical details; Justin D. Strait for helpful reading and suggestions. HL was supported by the Director, Office of Science, of the U.S. Department of Energy under Contract No. DE-AC02-05CH11231., who wants to thank the support of LBNL CRD during this research. DL wants to thank David Dunson and Tarek Zikry for motivating discussions. DL was supported by NIH/NCATS award UL1 TR002489, NIH/NHLBI award R01 HL149683 and NIH/NIEHS award P30 ES010126.

Our code for SRCA implementations and experiments are publicly available at \url{https://github.com/hrluo/SphericalRotationDimensionReduction}.
%\newpage
\bibliographystyle{chicago}
\bibliography{ref}

\newpage
\appendix
\section{\label{sec:orthogonalLoop}Orthogonal Loop Examples}
\begin{figure}[ht]
	\centering
	
	\centering\includegraphics[height=5cm]{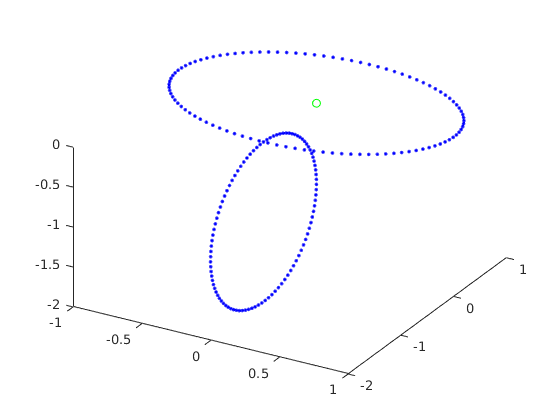}
	
	\caption{\label{fig_cc-pca_dataset_scatter} An example where PCA fails, adapted from \citet{luo_generalized_2020}. Note that the two orthogonal circles only intersect at one point. The DR results are summarized in Figure \ref{pca-spca_orthogonalLoops}}
	
	\label{fig:cc-pca_cc} 
\end{figure}

\begin{figure}
	\begin{center}
		\includegraphics[height=4cm]{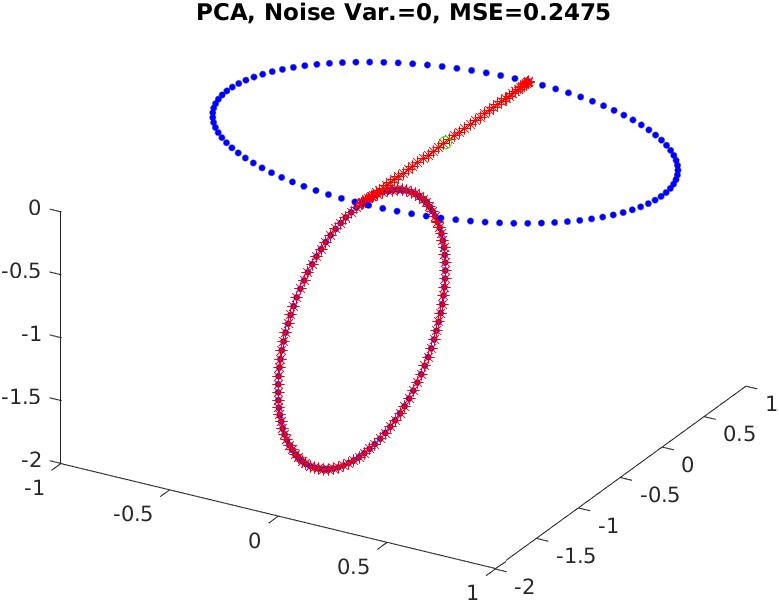}
		\includegraphics[height=4cm]{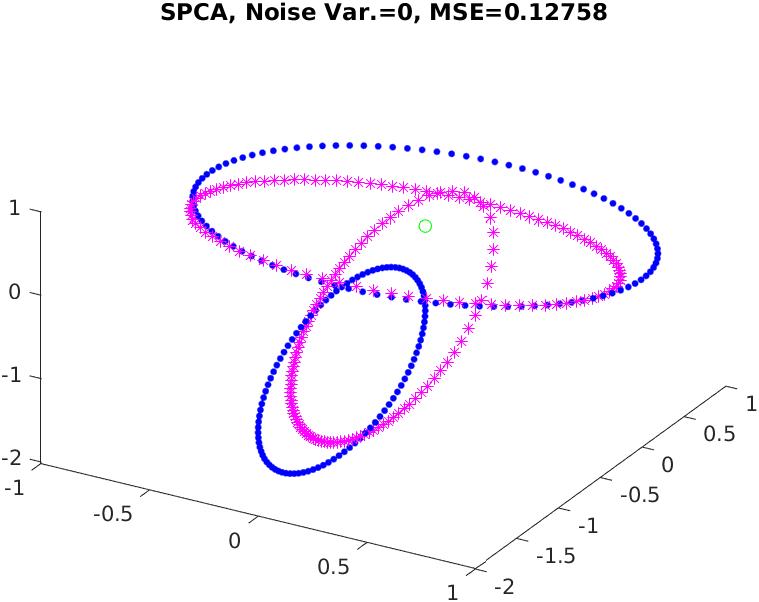}
		\includegraphics[height=4cm]{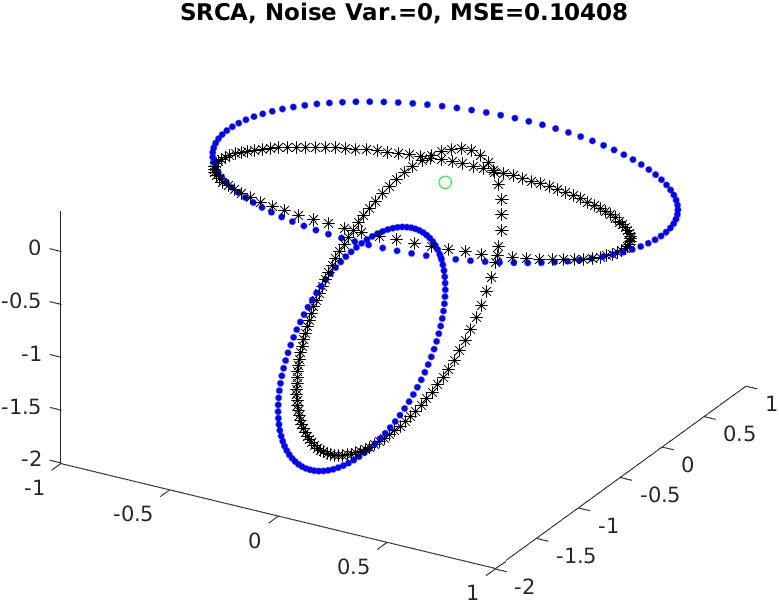}
	\end{center}
	\caption{The dimension-reduced dataset from the example shown in
		Figure \ref{fig:cc-pca_cc}. The green dot represents the origin $(0,0,0)$. The original dataset is represented by blue points. On the left panel, the PCA dimension-reduced dataset is represented by red stars.  On the middle and right panels the dimension-reduced
		dataset processed by SPCA and SRCA, is represented by magenta and black stars, respectively. 
	}\label{pca-spca_orthogonalLoops}
\end{figure}

Figure \ref{fig:cc-pca_cc} shows orthogonal circles parallel to $xz$ and $xy$ planes respectively in $\mathbb{R}^3$. The projection of these two circles to the principal axes given by PCA is shown in Figure \ref{pca-spca_orthogonalLoops}, where only one circular structure is retained in the reduced dataset while the other circular structure is completely destroyed %We consider the   dataset shown in Figure \ref{fig_cc-pca_dataset_scatter}.

\begin{figure}[h!]
	\centering
	\begin{adjustbox}{width=\textwidth}
		\begin{tabular}{ccc}
			PCA & SPCA & SRCA \tabularnewline
			\midrule
			\includegraphics[height=5cm]{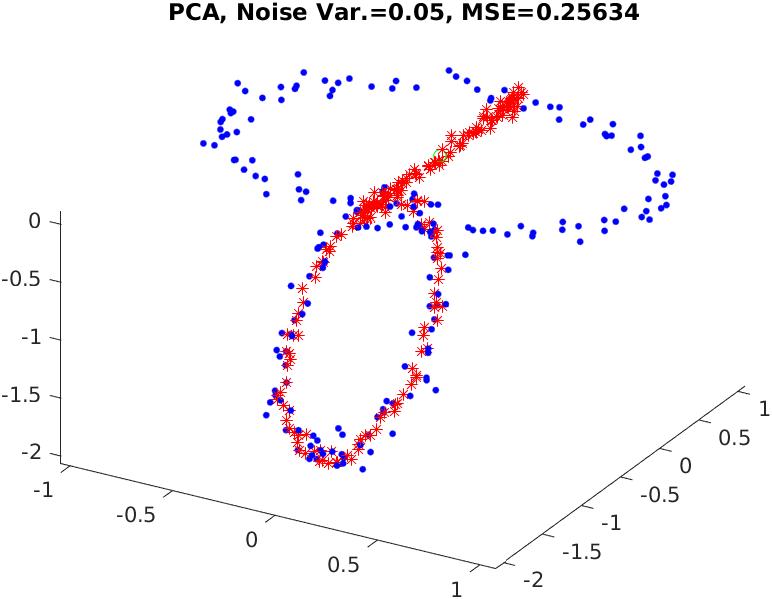}  & \includegraphics[height=5cm]{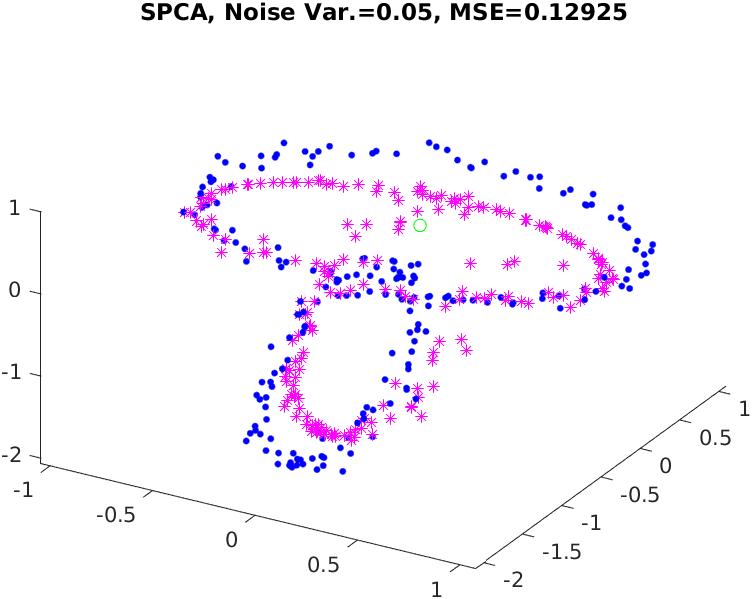} & \includegraphics[height=5cm]{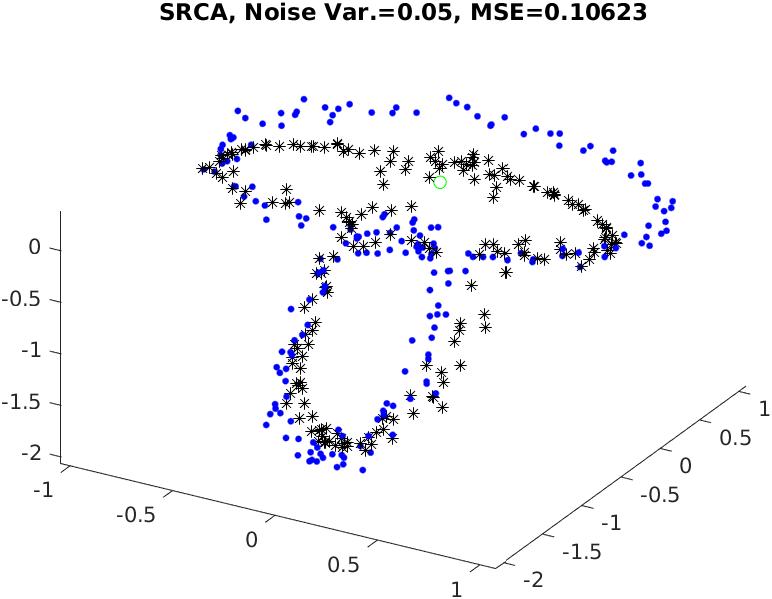} 
			\tabularnewline
			\includegraphics[height=5cm]{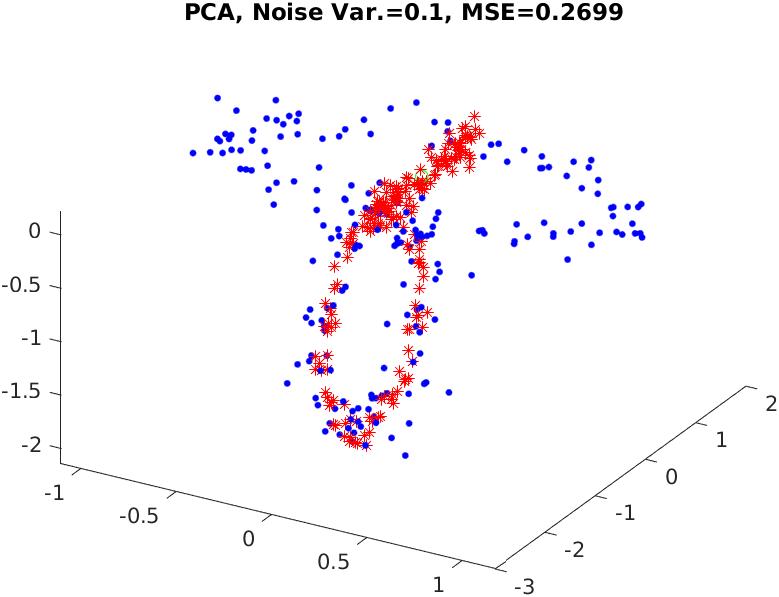}  & \includegraphics[height=5cm]{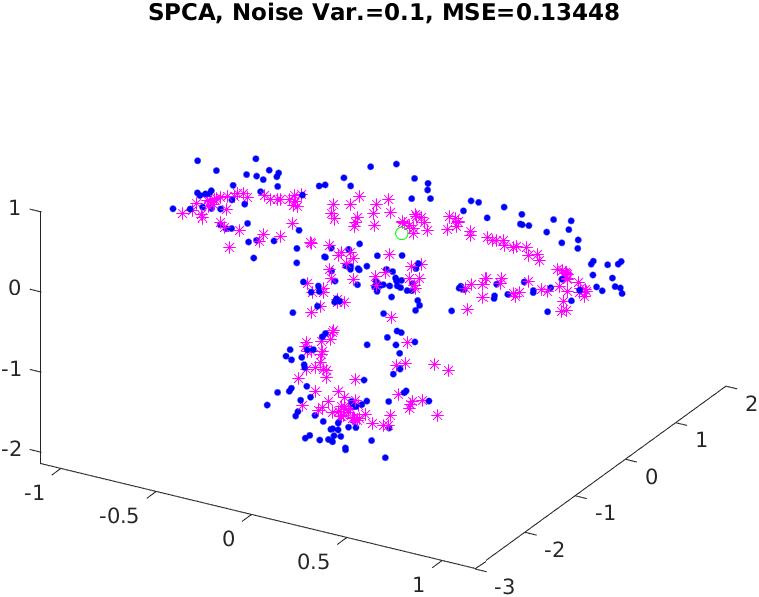} & \includegraphics[height=5cm]{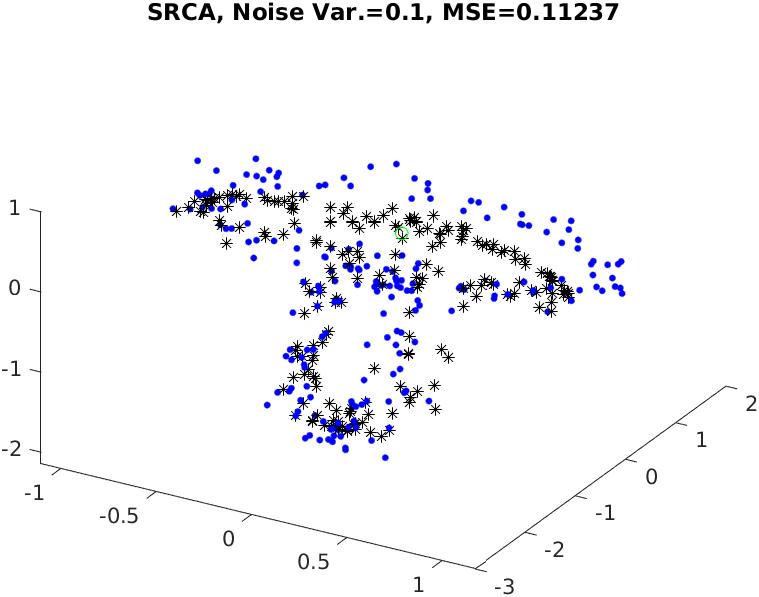} 
			\tabularnewline
			\includegraphics[height=5cm]{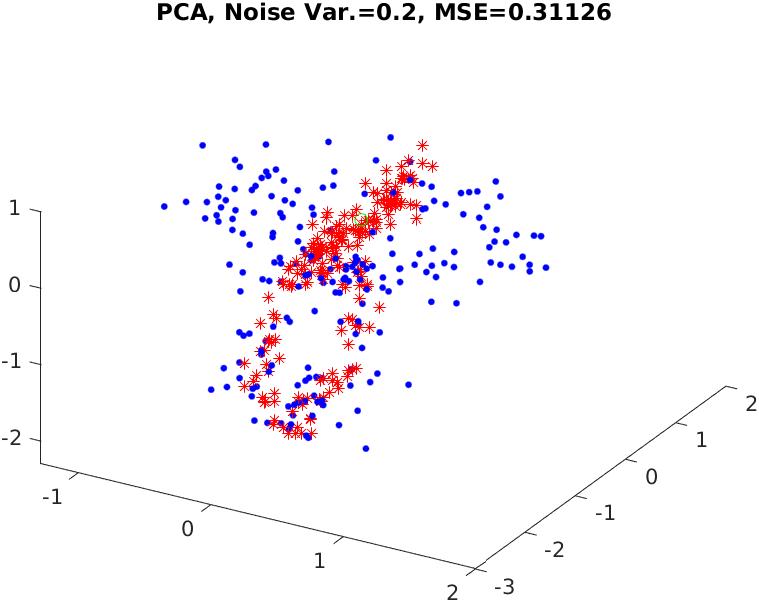}  & \includegraphics[height=5cm]{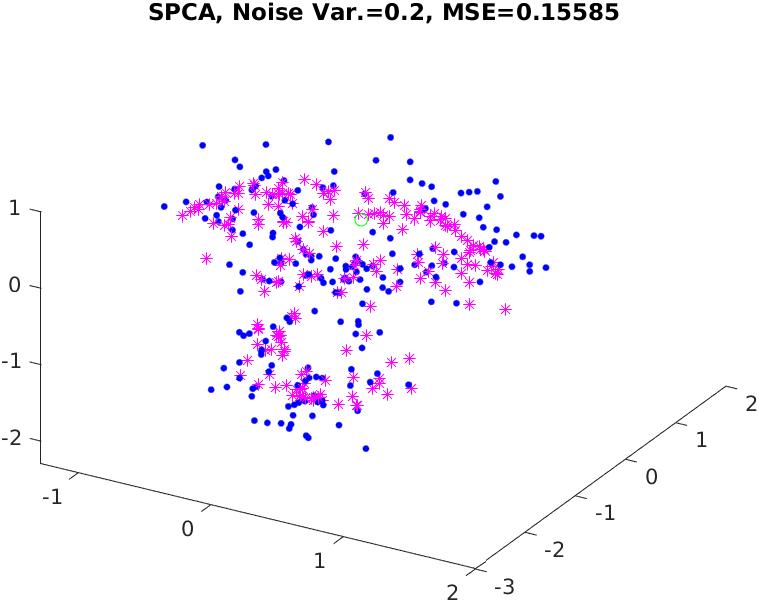} & \includegraphics[height=5cm]{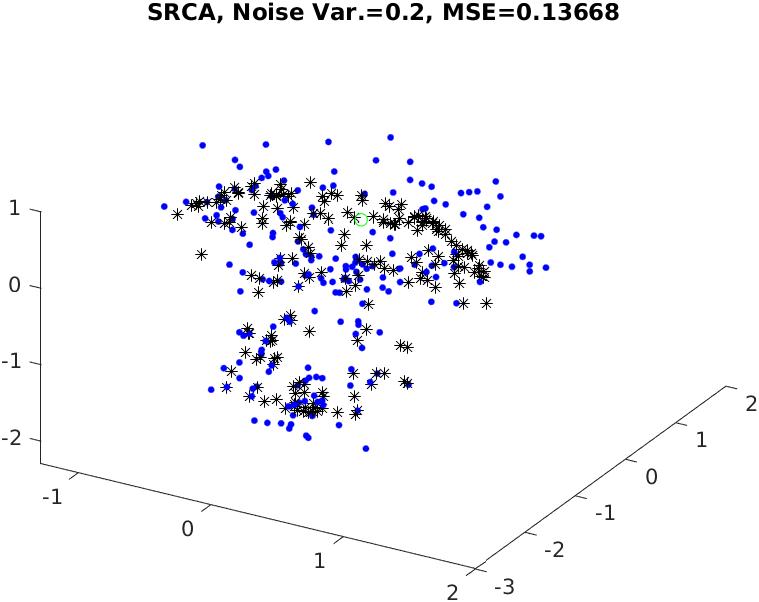} 
			\tabularnewline
		\end{tabular}
	\end{adjustbox}
	\caption{\label{fig:Example-OrthogonalLoop-noise}
		On the left columns we show the original dataset in blue points, and
		the PCA dimension-reduced dataset in red stars.  
		On the middle column we show the original dataset in blue points, and
		the SPCA dimension-reduced dataset in magenta stars. 
		On the right column we show the original dataset in blue points, and
		the SRCA dimension-reduced dataset in black stars. }
\end{figure}

In Figure \ref{fig:Example-OrthogonalLoop-noise}, we have the same  but each coordinates is perturbed by a Gaussian noise with mean zero and different noise variances. As the noise variance increases, we observe that the topological structure of this example of two orthogonal loops becomes less and less obvious. We can see that SRCA is consistently achieving the lowest matched  MSE defined in Section \ref{sec:MatchedDistMSE}, while both SPCA and SRCA preserves the topological structure relatively well. 
It becomes evident that that SPCA and SRCA method better respect the topology of the original dataset under the same $d'$. PCA does not retain the circular structure, but SRCA puts both circles onto a larger 1-sphere congruent to $\mathbb{S}^1$.

In Table \ref{table:OrthogonalLoop_MSE_noise}, we provide MSE for more settings of noise variances to show the MSE from each different DR methods. It can be observed that PCA becomes worse quickly in terms of MSE.

\begin{table}[h!]
	\begin{centering}
		\begin{tabular}{cccccccc}
			\hline 
			Noise Var. & 0 & 0.01 & 0.05 & 0.10 & 0.20 & 0.40 & 1.00\tabularnewline
			\hline 
			PCA  & 0.24750 & 0.24889  & 0.25634 & 0.26990  & 0.31126  & 0.45045 & 1.2653\tabularnewline
			\hline 
			SRCA & 0.10408 & 0.10421 & 0.10623 & 0.11237 & 0.13668  & 0.21834 & 0.64711\tabularnewline
			\hline 
			SPCA & 0.12758  & 0.12764  & 0.12925 & 0.13448  & 0.15585  & 0.22861 & 0.65268\tabularnewline
			\hline 
		\end{tabular}
		\par\end{centering}
	\caption{\label{table:OrthogonalLoop_MSE_noise}MSE for different DR methods performed on the same orthogonal loop
		dataset but with different noise variances in the Gaussian perturbation.}
\end{table}
\FloatBarrier

\section{\label{sec:basic-example}Other Synthetic  Examples}
\begin{figure}[h!]
	\begin{centering}
		
		\begin{tabular}{cc}
			%Trule 
			\hline 
			(a) Plane & (b) Torus\tabularnewline
			\hline 
			\includegraphics[scale=0.5]{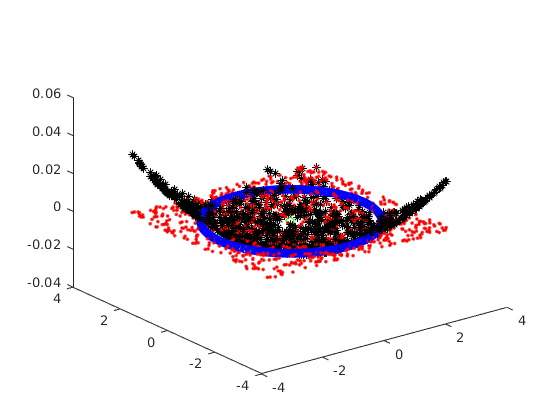} & \includegraphics[scale=0.5]{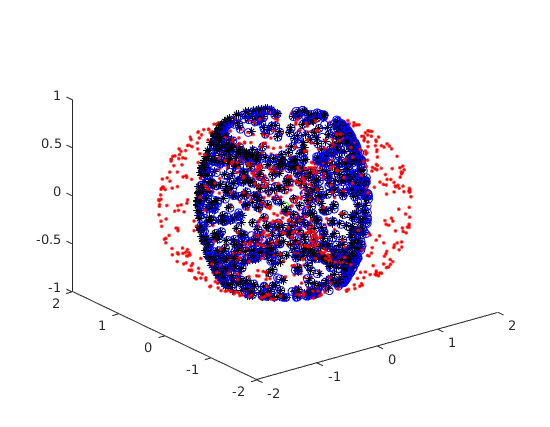}\tabularnewline
			\hline 
			
			(c) Sphere & (d) GEM\tabularnewline
			\hline 
			\includegraphics[scale=0.5]{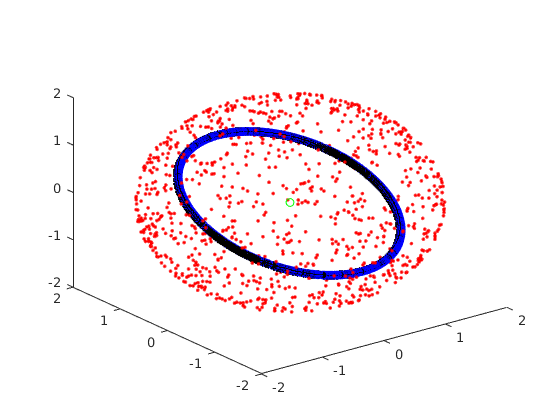} & \includegraphics[scale=0.5]{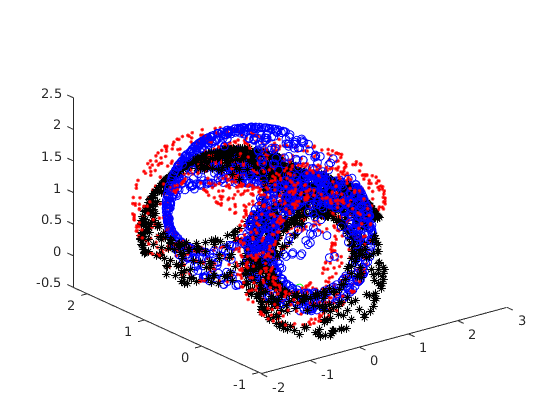}\tabularnewline
			\hline 
		\end{tabular}
		
	\end{centering}
	
	\caption{The datasets for the basic example. In these examples the datasets
		are observed in $\mathbb{R}^{3}$ ($d=3$) and we want to reduce the
		dataset by one dimension ($d'=2$). In each of the above panels, the
		red solid dots are the original datasets sampled from a uniform distribution
		lying on (a) plane (b) torus (c) sphere (d) triple torus without interior
		intersection (GEM). The blue circles are the points in reduced dataset
		obtained by SPCA. The black stars are the points in reduced dataset
		obtained by SRCA. We do not display the result of PCA in these figures. }
\end{figure}

In this appendix, we consider several basic examples we use to illustrate
the difference between PCA, SRCA, and SPCA. Below, we provide the quantitative measures
for each of these examples and the coranking summaries. In (a) plane,
we uniformly sample points $(x_{1},x_{2},0)$ from $[-3,3]\times[-3,3]\times\{0\}$.
In (b) torus, we take the parameterization $$\left((R_{1}+R_{2}\cos\theta)\cos\phi,(R_{1}+R_{2}\cos\theta)\sin\phi,R_{2}\sin\theta\right)$$
where $R_{1}=1/2$ and $R_{2}=1/3$. The parameters $(\theta,\phi)$
are uniformly sampled from $[0,2\pi)\times[0,2\pi)$. In (c) sphere,
we take the canonical parameterization and uniform sampling on the
parameter space $[0,2\pi)\times[0,2\pi)$, which is equivalent to
von-Mises Fisher distribution with concentration $\kappa=0$. In (d)
triple torus, we independently sampled 3 batches of points with equal sample sizes. Then we multiply each of these 3
batches with the following rotation matrices and add a translation vector:
\begin{align*}
	{R}_{1}=\left(\begin{array}{ccc}
		1 & 0 & 0\\
		0 & \cos\frac{\pi}{2} & -\sin\frac{\pi}{2}\\
		0 & \sin\frac{\pi}{2} & \cos\frac{\pi}{2}
	\end{array}\right), & {\tau}_{1}=\left(\begin{array}{c}
		0\\
		0\\
		3
	\end{array}\right).\\
	{R}_{2}=\left(\begin{array}{ccc}
		\cos\frac{\pi}{4} & 0 & \sin\frac{\pi}{4}\\
		0 & 1 & 0\\
		-\sin\frac{\pi}{4} & 0 & \cos\frac{\pi}{4}
	\end{array}\right), & {\tau}_{2}=\left(\begin{array}{c}
		0\\
		3\\
		3
	\end{array}\right).\\
	{R}_{3}=\left(\begin{array}{ccc}
		\cos0 & -\sin0 & 0\\
		\sin0 & \cos0 & 0\\
		0 & 0 & 1
	\end{array}\right), & {\tau}_{3}=\left(\begin{array}{c}
		3\\
		3\\
		3
	\end{array}\right).
\end{align*}
The resulting dataset contains three tori, and it is not difficult
to verify that these three tori do not have interior intersections.
We scale the dataset by subtracting $(1,1,1)^{T}$ from each point
and multiply $1/2$ entry-wisely. 

\begin{table}[h!]
	\begin{centering}
		
		\begin{tabular}{ccccccc}
			\hline 
			& SRCA & SPCA & PCA & LLE & tSNE & UMAP\tabularnewline
			\hline 
			CC & 0.9999998 & 0.9998008 & 1.0000000 & 0.9983610 & 0.9774094 & 0.9906862\tabularnewline
			\hline 
			AUC & 0.9995782 & 0.9898617 & 0.9998370 & 0.9604653 & 0.8670743 & 0.9094597\tabularnewline
			\hline 
			WAUC & 0.9990011 & 0.9913487 & 0.9991016 & 0.9725947 & 0.8460439 & 0.8320849\tabularnewline
			\hline 
		\end{tabular}\label{table:Plane_coranking}
		
	\end{centering}\caption{Performance scores and the coranking of plane.}
\end{table}

\begin{table}[h!]
	\begin{centering}
		
		\begin{tabular}{ccccccc}
			\hline 
			& SRCA & SPCA & PCA & LLE & tSNE & UMAP\tabularnewline
			\hline 
			CC & 0.8222110 & 0.8208433 & 0.9777033 & 0.9751233 & 0.8523848 & 0.9072153\tabularnewline
			\hline 
			AUC & 0.6217176 & 0.6194763 & 0.8317488 & 0.8251250 & 0.6344541 & 0.7257927\tabularnewline
			\hline 
			WAUC & 0.6025524 & 0.6033849 & 0.6104916 & 0.6092178 & 0.7430941 & 0.6707993\tabularnewline
			\hline 
		\end{tabular}\label{table:Torus_coranking}
		
		\caption{Performance scores and the coranking of torus.}
	\end{centering}
\end{table}

\begin{table}[h!]
	\begin{centering}
		
		\begin{tabular}{ccccccc}
			\hline 
			& SRCA & SPCA & PCA & LLE & tSNE & UMAP\tabularnewline
			\hline 
			CC & 1.0000000 & 1.0000000 & 0.8026803 & 0.7262926 & 0.6762080 & 0.6810310\tabularnewline
			\hline 
			AUC & 0.9999992 & 1.0000000 & 0.5585561 & 0.4876606 & 0.5171233 & 0.5696564\tabularnewline
			\hline 
			WAUC & 0.9999998 & 1.0000000 & 0.4953096 & 0.4792300 & 0.7312668 & 0.7270227\tabularnewline
			\hline 
		\end{tabular}\label{table:Sphere_coranking}

	\end{centering}
	
	\caption{Performance scores and the coranking of sphere.}
\end{table}

\begin{table}[h!]
	\begin{centering}
		
		\begin{tabular}{ccccccc}
			\hline 
			& SRCA & SPCA & PCA & LLE & tSNE & UMAP\tabularnewline
			\hline 
			CC & 0.9986899 & 0.7407685 & 0.9987456 & 0.9963411 & 0.5697951 & 0.9438497\tabularnewline
			\hline 
			AUC & 0.9509051 & 0.5618742 & 0.9518870 & 0.9205247 & 0.4646542 & 0.8042776\tabularnewline
			\hline 
			WAUC & 0.7067388 & 0.5204686 & 0.7057702 & 0.6973970 & 0.7077429 & 0.6735852\tabularnewline
			\hline 
		\end{tabular}\label{table:TripleTorus_coranking}
		
	\end{centering}\caption{Performance scores and the coranking of GEM.}
\end{table}
\FloatBarrier

From the performance evaluation table above, we can see that:
\begin{enumerate}
	\item In the plane example, all linear and non-linear DR methods performs
	well in terms of performance evaluation measures and coranking summaries. 
	\item In the torus example, neither SPCA nor SRCA perform as good as
	other DR methods in terms of scores. However, in the visualization
	of the reduced dataset, we can see that both SPCA and SRCA preserve
	the structure of torus pretty well.
	\item In the sphere example, both SPCA and SRCA give almost identical
	results, outperforming both the simplest PCA and the more sophisticated
	non-linear DR methods like tSNE and UMAP. In the visualization of
	the reduced dataset, we can see that both SPCA and SRCA reduce
	the data points on the sphere ($d=3$) to points on a geodesic circle
	$(d'=2)$. This preserves the spherical nature of the dataset and
	yield a better result. 
	\item In the GEM example, the SRCA is the best in terms of almost
	all performance scores. A visual verification also reveals that
	the resulting reduced dataset preserves all three holes in the tori. 
\end{enumerate}

\section{\label{sec:SPCA-Algorithm}SRCA Algorithms}

In this section, we present the algorithm that solves our optimization
problem (\ref{eq:SPCA_standard}) and (\ref{eq:SPCA_L1}). Algorithm
\ref{alg:SPCA} delineates the binary search strategy which exhausts $2^{d}$
possible subsets of $\{1,\cdots,d\}$ to find an optimal subspace.
Algorithm \ref{alg:SPCA_L1} delineates the $l_{1}$-relaxation strategy
which formulates the original problem (\ref{eq:SPCA_standard}) into
an optimization problem (\ref{eq:SPCA_L1}) to find an  optimal
subspace. 

Detailed explanation of algorithms in this section is as follows. 

\begin{enumerate}%[leftmargin=*]
	\item (Step 1: Get the empirical mean for $\mathcal{X}$.) Estimate the
	empirical mean $\bar{\mathcal{X}}$ for the dataset $\mathcal{X}$
	in $\mathbb{R}^{d}$, and subtract the mean $\bar{\mathcal{X}}$ to
	make sure that the assumption of PCA is satisfied. 
	\begin{align*}
		{z}_{i} & ={x}_{i}-\frac{1}{n}\sum_{i=1}^{n}{x}_{i} = {x}_i-\bar{{x}}
	\end{align*}
	\item (Step 2: Conduct the rotation.) We choose a rotation method
	to construct rotation matrix ${R}$ based on the datset $\mathcal{\ensuremath{X}}$.
	Then we rotate the dataset $\mathcal{X}$ to standard position $(\mathcal{X}-\bar{\mathcal{X}}){R}$.
	Here we use a chosen rotation matrix ${R}$ (PCA, ICA or other kinds of optimal
	rotation) to rotate the
	sphere so that all its axes are parallel to the coordinate
	axes. \\
	%The optimal rotation will bring the most variate directions to be parallel to the coordinate axes. 
	For PCA rotation, let the covariance matrix $\mathrm{cov}({z}_{i})={R}{\Lambda}{R}^{T}$
	where ${\Lambda}$ is diagonal and the rotation matrix ${R}$ is orthogonal.
	\begin{align*}
		{y}_{i} & ={R}{z}_{i}
	\end{align*}
	\item (Step 3: Binary search for the best $d'+1$ axes.) In this step we perform
	dimension reduction. 
	Now that we can assume that the axes of the sphere (or ellipsoid)
	are parallel to the coordinate axes. We solve the binary optimization
	problem (\ref{eq:SPCA_standard}) with ${W}={I}$ (or other
	${W}$ if extremely skewed dataset is observed) to choose the optimal
	directions to retain. In this step we would find the optimal ${v}_{\text{opt}}$ and hence the optimal index
	set $\mathcal{I}_{\text{opt}}$. \\
	In this optimization problem, as we stated in (\ref{eq:SPCA_standard})
	above, we conduct dimension reduction by minimizing the loss function
	based on the point-to-ellipsoid distance to the estimated sphere $S_{\mathcal{I}}$.
	
	The optimization problem is spelled out as (\ref{eq:SPCA_standard-1}).
	\item (Step 4: Eliminate the un-chosen dimensions.) This simply sets drop the
	un-selected dimension not in $\mathcal{I}_{\text{opt}}$, equivalently, we find the ${v}_{\text{opt}}$ in the notation
	of problems (\ref{eq:SPCA_standard}) and (\ref{eq:SPCA_L1}). 
	\begin{enumerate}
		\item In the standard form (\ref{eq:SPCA_standard}), we may let ${\eta}={v}_{\text{opt}}$
		be the binary vector such that $\|{v}_{\text{opt}}\|_{l_{0}}=d'+1$. 
		\item In the $l_{1}$-relaxed form (\ref{eq:SPCA_L1}), the ${v}_{\text{opt}}$
		would have $l_{1}$-norm less or equal than $d'+1$ but not binary
		entries. We construct a binary vector ${\eta}$ such that only
		the first leading $d'+1$ entries with the largest absolute values in
		${v}_{\text{opt}}$ are 1; and the rest entries are 0.
	\end{enumerate}
	\item (Step 5: Re-estimate the center and radius.) Only in the problem (\ref{eq:SPCA_L1}),
	we re-estimate the center ${c}_{\text{opt}}$ and radius $r_{\text{opt}}$
	using the same loss function but a fixed ${v}_{\text{opt}}$. In
	standard problem (\ref{eq:SPCA_standard}), we use the center ${c}_{\text{opt}}$
	and radius $r_{\text{opt}}$ from step 3.
	\item (Step 6: Project and rotate the sphere back into full space.) After
	we choose the dimension and axes, we project the dataset onto a sphere
	with the center ${c}_{\text{opt}}$ and radius $r_{\text{opt}}$
	and add back the empirical mean $\bar{\mathcal{X}}$. More specifically,
	we project the datapoints ${x}_{i}$ to the sphere $S({c},r)$ as in (\ref{eq:project-to-sphere}) and then we simply we rotate the resulting dimension-reduced dataset back,
	using the inverse of the same rotation matrix ${R}$. 
\end{enumerate}

{\singlespacing
\LinesNumberedHidden
\setcounter{algoline}{0}
\begin{adjustbox}{width=1.\textwidth,center}
	\begin{algorithm}[H]
		\caption{\label{alg:SPCA_L1} SRCA dimension reduction algorithm with $l_1$ relaxation} 
		\KwData{$X$ (data matrix consisting of $n$ samples in $\mathbb{R}^d$)} 
		\KwIn{$d'$ (the dimension of the sphere), ${W}$ (the covariance weight matrix, by default ${W}={I}_d$), $\lambda$ (optional, the sparse penalty parameter), rotationMethod (the method we use to construct the rotation matrix).}
		\KwResult{
			$\hat{{c}}$ (The estimated center of $S_{\mathcal{I}}$ in $\mathbb{R}^d$), $\hat{r}$ (The estimated radius of $S_{\mathcal{I}}$), $\mathcal{I}_{opt}$ (The optimal index subset) 
		}
		\SetKwFunction{GetRotation}{GetRotation} \GetRotation($X$,rotationMethod) obtains a $d\times d$ rotation matrix based on the data matrix $X$. The rotationMethod option specifies what method we use to construct the rotation matrix, by default, we use PCA to obtain the rotation matrix. \\
		\SetKwFunction{ProjectToSphere}{ProjectToSphere} \ProjectToSphere($X$,${c}$,$r$,$k$) is a projection that projects the point set $X$ onto an $l_k$-sphere of center ${c}$ and radius $r$ via $X\mapsto{c}+\frac{X-{c}}{\|X-{c}\|_{l_k}}\cdot r$.\\
		
		\Numberline \Begin { 
			\Numberline Standardize the dataset by subtracting its empirical mean $X$ = $X-\bar{X}$ \\
			\Numberline Construct a rotate matrix $R$ = \GetRotation($X$,rotationMethod) \\     
			\Numberline $X_{rotated} = X* R$ \\
			
			\Numberline Solve the optimization problem (\ref{eq:SPCA_L1}) with respect to ${c},r$ and ${v}$\\
			\Numberline Denote the solution as ${c}_{0},r_{0},{v}_{opt}$ \\
			
			\Numberline Construct $\mathcal{I}$ that contains the largest/leading $d'+1$ coordinates of ${v}_{opt}$. \\
			
			\Numberline Solve the optimization problem (\ref{eq:SPCA_standard}) with respect to ${c},r$ with a fixed $\mathcal{I}$\\
			\Numberline Denote the solution as ${c}_{opt},r_{opt}$ \\
			\Numberline $\hat{{c}} = {c}_{opt}\cdot{\eta}*R^{-1}+\bar{X}$ \\
			\Numberline $\hat{r} = r_{opt}$ \\
			\Numberline $X_{rotated}(:,\mathcal{I}) \leftarrow {0}$ \\
			\Numberline $X_{rotated}\leftarrow$ \ProjectToSphere($X$,$\hat{{c}}$,$\hat{r}$,$k$) \\
			\Numberline $X_{output} \leftarrow X_{rotated}*R^{-1} + \bar{X}$ \\
		}
	\end{algorithm}
\end{adjustbox}
\FloatBarrier
}
\section{Dataset Selection}\label{sec: Dataset Selection}

We select datasets for high- and low-dimensional scenarios, covering both $n\geq d$ and $n<d$ as detailed below. Thanks to the binary search scheme we designed in our
SRCA algorithm, SRCA can be parallelized when an even larger $n$
(e.g., $n=100,000$) presents. Neither PCA nor non-linear methods we consider below can be generalized to a quite large $n$ in an obvious way. 

The tSNE \citep{van2008visualizing} has known problems of not distance-preserving and creating false clusters in the dimension-reduced dataset \citep{schubert2017intrinsic}. The UMAP \citep{mcinnes2018umap-software} has known problems of being sensitive to outliers and require the user to have a good understanding of their distance metrics to interpret the dimension-reduced dataset. To highlight the advantage of our method with a geometric loss function, we also choose labeled datasets to study the structure-preserving properties and datasets that require careful normalization. 

We use several public datasets for our numerical experiments, but consider only datasets with continuous variables as its features,
as we recognized that dimension reduction for datasets with discrete (categorical and integer)
or mixed type variables as their attributes is a different problem  \citep{scholkopf1997kernel,scholkopf1998nonlinear}.

Before the analysis of our experiments,
we shall briefly introduce our datasets:
\begin{itemize}%[leftmargin=*]
	\item Source
	\begin{itemize}
		\item UCI repository (\url{https://archive.ics.uci.edu/ml}): Banknote, Climate, Concrete, Ecoli\footnote{Three groups among them with sample size smaller than 5 are removed for visual convenience. The five groups left are cytoplasmic proteins (cp), inner membrane
			proteins without a signal sequence (im), inner brane proteins with an uncleavable signal sequence (imU), other outer membrane proteins (om) and periplasmic proteins (pp).}, Leaf, PowerPlant, UserKnowledge.
		\item Microarray:  Alon \citep{Alon:1999dy}. 
		\item GTEx (\url{https://gtexportal.org/home/}).
	\end{itemize}
	\item Sample size\\
	We understand that a valid dimension reduction method should have
	reasonable performance regardless of the size of the underlying datasets.
	We choose a wide range of datasets with sample sizes varying from
	100 to 10,000. In the table below, we show the code for each dataset with its
	sample size and dimension $(n\times d)$.
	
	\begin{adjustbox}{center}%
		\begin{tabular}{cc}%
			$n>d$ & $n\leq d$\tabularnewline
			\midrule 
			Banknote ($1372\times4$), UserKnowledge ($403\times5$), & Kidney Medulla ($4\times 500$),
			\tabularnewline
			Ecoli ($336\times7$), Concrete ($1030\times8$), & Fallopian Tube ($9\times 500$), \tabularnewline
			Climate ($540\times18$),  Leaf ($340\times14$),  
			%Sorlie ($85\times456$), CNAE9 ($1080\times856$),  
			& Cervix Endocervix ($10\times 500$),.\tabularnewline
			PowerPlant ($9568\times5$), & Alon ($62\times2000$),
			.%, 3newsgroup ($300\times12679$). 
			\tabularnewline
			\bottomrule
	\end{tabular}\end{adjustbox}\\
	
	\item Dimensionality\\
	It is of central importance to recognize that the relation between
	the sample size $n$ and the original dimension $d$ would affect
	dimension reduction methods. In fact, the sparse PCA \citep{erichson2020sparse}
	were developed to take the sparsity (i.e., $n<d$) of the dataset
	into consideration. SRCA has a natural sparsity penalty parameter
	in the loss function $\mathscr{L}$ we designed. To see the performance
	of different methods on dense ($n>d$) and sparse ($n\leq d$) data, we
	also include both kinds of datasets in the above selection. 
	\item Normalization\\
	When the attributes or features of the dataset have high mutual correlation
	or relationships with one another (e.g., total expenditure cannot exceed
	total income of an individual), normalization would introduce problems
	like distortion of correlation and violation of relationships. When
	the attributes or features of the dataset are uncorrelated or independent,
	normalization would convert all features to (relatively) the same
	scale. We include datasets that require normalization and those that
	do not require normalization. 
	\begin{itemize}
		\item Not normalized: Banknote, Ecoli, PowerPlant, UserKnowledge, Climate.%Sorlie,
		
		\item Normalized:  Alon, Concrete, Leaf, GTEx.% 3newsgroup (tf-idf),CNAE9 (tf-idf),
	\end{itemize}
\end{itemize}

Finally, we shall point out that we have not covered 
any dataset with a high $d$ and large $n$. Our exploratory experiments
found that the performance of existing dimension methods, including
modern methods, has suffered from a very high computational cost. It is a separate problem to study how to perform dimension reduction on a large high-dimensional dataset. 

Decomposition-based methods like PCA, multi-dimensional scaling (MDS) and truncated singular value decomposition (SVD) become
exceedingly slow for a dataset with large $n$ and $d$.
Manifold learning based methods like tSNE, UMAP and IsoMap \citep{tenenbaum2000global} have some
variation in computational time due to their stochastic nature, but
they are all rather slow. Therefore, we would leave this type of dataset
as a separate problem that we do not experiment in the current paper. 

\section{\label{sec: Coranking Tables} Coranking Performance Comparison for Section \ref{sec: Coranking matrix}}

\begin{table}[h!]
	\centering 
	\global\long\def\~{\hphantom{0}}%

	\begin{tabular}{ccccccc}
		\hline 
		& SRCA  & SPCA  & PCA & LLE & tSNE & UMAP\tabularnewline
		\hline 
		CC  &0.987& 0.925& 0.988& 0.833 &0.640& 0.635\tabularnewline
		\hline 
		AUC & 0.869 &0.774 &0.860& 0.598 &0.469& 0.459\tabularnewline
		\hline
		WAUC  & 0.644 &0.582&0.626& 0.503 &0.694& 0.600\tabularnewline
		\hline 
	\end{tabular} \caption{Coranking performance scores of Banknote}\label{table:Banknote_coranking} 
\end{table}

\begin{table}[h!]
	\centering 
	\global\long\def\~{\hphantom{0}}%
	
	\begin{tabular}{ccccccc}
		\hline 
		& SRCA  & SPCA  & PCA & LLE & tSNE & UMAP \tabularnewline
		\hline 
		CC  &0.270& 0.270 &0.262 &0.464& 0.215& 0.124\tabularnewline
		\hline 
		AUC  & 0.152 &0.152& 0.148& 0.225 &0.147& 0.102\tabularnewline
		\hline 
		WAUC  & 0.134 &0.132 &0.0864& 0.105 &0.141& 0.118\tabularnewline
		\hline 
	\end{tabular} \caption{Coranking performance scores of Ecoli}\label{table:Ecoli_coranking} 
\end{table}

\begin{table}[h!]
	\centering 
	\global\long\def\~{\hphantom{0}}%
	
	\begin{tabular}{ccccccc}
		\hline 
		& SRCA  & SPCA  & PCA & LLE & tSNE & UMAP \tabularnewline
		\hline 
		CC  &0.987& 0.815& 0.987 &0.928& 0.620& 0.847\tabularnewline
		\hline 
		AUC  & 0.886 &0.605 &0.886& 0.731 &0.446& 0.651\tabularnewline
		\hline 
		WAUC  & 0.485 &0.416 &0.485& 0.447 &0.611& 0.528\tabularnewline
		\hline 
	\end{tabular} \caption{Coranking performance scores of PowerPlant}\label{table:Power_coranking} 
\end{table}

\iffalse
\begin{table}[h!]
	\centering 
	\global\long\def\~{\hphantom{0}}%
	
	\begin{tabular}{ccccccc}
		\hline 
		& SRCA  & SPCA  & PCA & LLE & tSNE & UMAP \tabularnewline
		\hline 
		CC  &0.937 &0.906 &0.933&0.685& 0.681& 0.754\tabularnewline
		\hline 
		AUC  & 0.745 &0.642 &0.789& 0.420& 0.490& 0.675\tabularnewline
		\hline 
		WAUC  & 0.502& 0.471& 0.502& 0.383&0.610& 0.533\tabularnewline
		\hline 
	\end{tabular} \caption{Coranking performance scores of GalaxyZoo}\label{table:Galaxy_coranking} 
\end{table}
\fi

\begin{table}[h!]
	\centering 
	\global\long\def\~{\hphantom{0}}%
	
	\begin{tabular}{cccccccc}
		\hline 
		& SRCA  & SPCA  & PCA & LLE & tSNE & UMAP \tabularnewline
		\hline 
		CC &0.219 &0.211 &0.219 &0.183& 0.134& 0.180\tabularnewline
		\hline 
		AUC  & 0.0780 &0.0853 &0.0785& 0.0700& 0.0707& 0.0635\tabularnewline
		\hline 
		WAUC  & 0.0935 &0.0952 &0.0948& 0.0762& 0.106& 0.100\tabularnewline
		\hline 
	\end{tabular} \caption{Coranking performance scores of Leaf}\label{table:Leaf_coranking} 
\end{table}

\begin{table}[h!]
	\centering 
	\global\long\def\~{\hphantom{0}}%
	
	\begin{tabular}{cccccccc}
		\hline 
		& SRCA  & SPCA  & PCA & LLE & tSNE & UMAP \tabularnewline
		\hline 
		CC  &0.730 &0.798& 0.693 &0.585& 0.483 &0.592  \tabularnewline
		\hline 
		AUC  &  0.416& 0.456 &0.379 &0.338& 0.231 &0.384\tabularnewline
		\hline 
		WAUC  & 0.256 &0.3233& 0.251& 0.212 &0.214& 0.328\tabularnewline
		\hline 
	\end{tabular} \caption{Coranking performance scores of  Alon. 
	}\label{table:Alon_coranking} 
\end{table}
\FloatBarrier

\section{Sparse Penalty}
\label{sec: sparse penalty}

\begin{table}[h!]
	\centering 
	\global\long\def\~{\hphantom{0}}%
	\begin{tabular}{ccccccccc}
		\hline 
		$\lambda$   & $10^{-1}$  & $10^{-2}$ & $10^{-3}$ & $10^{-4}$ & $10^{-5}$ & $0$\tabularnewline
		\hline 
		CC   &0.704& 0.727 &0.730& 0.730 &0.730 &0.730\tabularnewline
		\hline 
		AUC & 0.408 &0.416 &0.416& 0.416 &0.416&0.416\tabularnewline
		\hline 
		WAUC   &0.261& 0.256& 0.256 &0.256& 0.256&0.256\tabularnewline
		%\hline 
		%$Q_{local}$  
		%& 0.455 &0.445& 0.446& 0.446 &0.446&0.446\tabularnewline
		%\hline 
		%$Q_{global}$   & 0.862& 0.860& 0.860& 0.860
		%& 0.860 &0.860 \tabularnewline
		\hline 
	\end{tabular} 
	\caption{Coranking performance scores for different $\lambda$'s on the Alon dataset, $d'=2$.}\label{table:Alon_coranking_lambda} 
\end{table}
Here, we provide a new version of SRCA with sparse penalty, which only involves an additional penalty term in the loss function we designed. 
Recall that the objective loss
function with a weighted matrix $W$  in our method is 
\begin{align*}
	d({x}_{i},S_{\mathcal{I}}({c},r))^{2}=({x}_{i}-{c})^{T} {W}({x}_i-{c})+r^{2}-2r\sqrt{\left({x}_{i}-{c}\right)^{T}\sqrt{{W}}^{T}{I}_{\mathcal{I}}\sqrt{{W}}\left({x}_{i}-{c}\right)}
\end{align*}
which involves only the point-to-sphere distance from ${x}_{i}$ to
the estimated sphere surface $S_{\mathcal{I}}$ (based on dataset
$\mathcal{X}=\{{x}_{1},{x}_{2},\cdots,{x}_{n}\}$). One problem
we wish to address when there exists sparsity in the dataset in the
procedure of dimension reduction, is that we want the sparsity being
preserved. 

To be more precise, if ${x}_{i}$'s have most coordinates
zeros except for a few, then we want the reduced dataset $\hat{{x}}_{i}$
to have a similar property. This can be achieved by penalizing $\|{I}_{\mathcal{I}}({x}_{i}-c)\|_{1}$
in the optimization problem%\st{(\ref{eq:basic opt}) and (\ref{eq:W opt})}
, which encourages the estimated sphere so that the data are in the affine subspace centered at $c$ while parallel to the coordinate planes. %For example, the three $S^1$ in Figure \ref{beautiful_spheres} are preferred when $d'=1$.
This is different from the $l_{1}$ relaxation we propose above.
The $l_{1}$ relaxation we proposed above is an approximation to the
constraints we imposed on the binary optimization problem. Here, we
directly penalize the reduced coordinates. %Note that after the optimal rotation, sparsity would not be disturbed. 
The corresponding optimization problem is: 
\begin{align}
	\min_{{c}\in\mathbb{R}^{d},r\in\mathbb{R}^{+}} & \sum_{i=1}^{n}\left(({x}_{i}-{c})^{T} {W}({x}_i-{c})+r^{2}\right.\nonumber\\
	& \left. -2r\sqrt{\left({x}_{i}-{c}\right)^{T}\sqrt{{W}}^{T}{v}^{T}{I}{v}\sqrt{{W}}\left({x}_{i}-{c}\right)}\right) +\lambda\|{I}_{\mathcal{I}}(x_i-c)\|_{1},\label{eq:general opt} \\
	& \text{ s.t. }\|{v}\|_{l_{k}}\leq d'+1,\lambda>0,
\end{align}
with a tuning parameter $\lambda>0$. %(The constraint
$\|{v}\|_{l_{k}}\leq d'+1$ can be $\|{v}\|_{l_{k}}=d'+1$.
This feature of $l_1$ constraint allows us to perform dimension reduction in a high-dimensional input space with SRCA. We want to consider the penalty parameter $\lambda$ that controls the retained dimension $d'$ when $l_1$ approximation is in place as defined in \eqref{eq:general opt}. When a strict binary search like $l_0$ optimization in \eqref{eq:SPCA_standard-1} is used, the penalty is usually not needed. 
However, the caution we shall take here is that 
the selection of sparsity penalty parameter $\lambda$ should roughly be at the same magnitude as the loss function in order to function properly.

The tuning parameter $\lambda>0$ is part
of the objective function, instead of the constraints.
In some applications, the penalty term can also be replaced with $\lambda\sum_{i=1}^n\|{I}_{\mathcal{I}}{x}_{i}\|_{1}$. 
In experiments, we found that SRCA is not sensitive to $\lambda$. In very high-dimensional datasets (e.g., Alon (see~Supplement~\ref{sec: Dataset Selection}), GTEx), the choice of this parameter also affects the convergence speed in the execution of the optimization algorithm, a larger penalty parameter forces the numerical algorithm to converge slightly faster.  
Alternatively, we can also treat the choice of this parameter as a apriori tuning parameter of the loss function, whose values can be selected for different datasets using cross-validation. 

\section{\label{sec:Sphere-Estimation-with}Spherical Estimation}

The essence of SPCA \citep{li_efficient_2022} can
be summarized as a two-step procedure: 
\begin{enumerate}%[leftmargin=*]
	\item First, we utilize the principal component analysis (PCA) to find a
	subspace $V\subset\mathbb{R}^{d'+1}$ of retained dimension based on $\mathcal{X}$ and
	project $\mathcal{X}$ to $\hat{\mathcal{X}}$ in $V$. 
	\item Second, we perform a circular (or $d'$-dimensional spherical)
	regression\footnote{Unfortunately, although methods in circular regression could be extended
		to spheres of intrinsic dimensions greater than 1, the term ``circular
		regression'' instead of ``spherical regression'' is adopted.} with the projected image $\hat{\mathcal{X}}$ onto $V$. 
\end{enumerate}
By selecting the principal components given by PCA, we find a subspace 
$V$ and determine the dimension of the $S$. By fitting a circular
regression on a $d'$-dimensional sphere with the projected
dataset $\hat{\mathcal{X}}$, we determine the center ${c}$ and
radius $r$ of the spherical support.

Suppose the assumed sphere $S_{V}({c},r)$ is $d^{2}({x},{c})=r^{2}$,
whose dimensionality is determined by the PCA estimated linear subspace
$V$. Two typical loss functions for the estimation of ${c},r$
are: 
\begin{align*}
	\mathscr{L}(V,{c},r) & =\sum_{i=1}^{n}d^{2}({x}_{i},S_{V}({c},r))\\
	& \text{where }d^{2}\text{ can be chosen as geometric or algebraic loss}\\
	\text{geometric loss} & =\sum_{i=1}^{n}\left(\sqrt{\left({x}_{i}-{c}\right)^{T}\left({x}_{i}-{c}\right)}-r\right)^{2}\\
	\text{algebraic loss} & =\sum_{i=1}^{n}\left(\left({x}_{i}-{c}\right)^{T}\left({x}_{i}-{c}\right)-r^{2}\right)^{2}
\end{align*}
Following \citet{li_efficient_2022}, we first assume that $V$ is
determined (through PCA) and attempt to estimate the center ${c}$
and radius $r$ via a two-step gradient descent with both geometric
and algebraic loss functions. Through the procedure of taking derivation,
we observe and explain why an analytic solution for ${c}$ and
$r$ is impossible in this SPCA setup in the end and how SRCA
handles this problem. 

\subsection{Geometric Loss}

Let us calculate the geometric loss first, the algebraic loss is calculated
at the end. This function is a quadratic polynomial of radius parameter
$r>0$, $\mathscr{L}$ has a unique global (conditional) minimum in
$r$ if $\hat{r}>0$. When the ${c}$ is assumed fixed.

We calculate its gradient 
\begin{align*}
	\frac{\partial\mathscr{L}({c},r)}{\partial r} & =\sum_{i=1}^{n}\frac{\partial}{\partial r}d^{2}({x}_{i},S({c},r))\\
	& =\sum_{i=1}^{n}\frac{\partial}{\partial r}\left(\sqrt{\left({x}_{i}-{c}\right)^{T}\left({x}_{i}-{c}\right)}-r\right)^{2}\\
	& =\sum_{i=1}^{n}-2\left(\sqrt{\left({x}_{i}-{c}\right)^{T}\left({x}_{i}-{c}\right)}-r\right)
\end{align*}
Setting this equation to zero, we have 
\[
\hat{r}=\frac{1}{n}\sum_{j=1}^{n}\sqrt{\left({x}_{j}-{c}\right)^{T}\left({x}_{j}-{c}\right)}\geq0.
\]
Plug this back into the $\mathscr{L}({c},r)$ we have 
\begin{align*}
	\mathscr{L}({c},\hat{r}) & =\sum_{i=1}^{n}d^{2}({x}_{i},S({c},\hat{r}))\\
	& =\sum_{i=1}^{n}\left(\sqrt{\left({x}_{i}-{c}\right)^{T}\left({x}_{i}-{c}\right)}-\hat{r}\right)^{2}\\
	& =\sum_{i=1}^{n}\left(\sqrt{\left({x}_{i}-{c}\right)^{T}\left({x}_{i}-{c}\right)}-\frac{1}{n}\sum_{j=1}^{n}\sqrt{\left({x}_{j}-{c}\right)^{T}\left({x}_{j}-{c}\right)}\right)^{2}
\end{align*}
Although this cannot be simplified further (due to the fact that it
is fourth power in ${c}$), we can still attempt to take its gradient

\begin{align*}
	\frac{\partial\mathscr{L}({c},\hat{r})}{\partial{c}} & =\sum_{i=1}^{n}\frac{\partial}{\partial{c}}\left(\sqrt{\left({x}_{i}-{c}\right)^{T}\left({x}_{i}-{c}\right)}-\frac{1}{n}\sum_{j=1}^{n}\sqrt{\left({x}_{j}-{c}\right)^{T}\left({x}_{j}-{c}\right)}\right)^{2},\\
	& \text{ where }\hat{r}=\frac{1}{n}\sum_{j=1}^{n}\sqrt{\left({x}_{j}-{c}\right)^{T}\left({x}_{j}-{c}\right)}\\
	& =\sum_{i=1}^{n}2\left(\sqrt{\left({x}_{i}-{c}\right)^{T}\left({x}_{i}-{c}\right)}-\hat{r}\right)\cdot\\
	& \left(\frac{\partial}{\partial{c}}\sqrt{\left({x}_{i}-{c}\right)^{T}\left({x}_{i}-{c}\right)}-\frac{1}{n}\sum_{j=1}^{n}\frac{\partial}{\partial{c}}\sqrt{\left({x}_{j}-{c}\right)^{T}\left({x}_{j}-{c}\right)}\right)
\end{align*}
The equation $\frac{\partial\mathscr{L}({c},\hat{r})}{\partial{c}}=0$
would not have an analytic solution in general. However, with an appropriate
gradient-based optimization method, for example, Gauss-Newton method
with Levenberg-Marquardt correction \citep{chernov_circular_2010},
the sequence of estimates of ${c},r$ can be proven to converge
to global minimum under the regularity condition. It is also not hard
to observe why the insertion of ${W}$ into the $\sqrt{\left({x}_{i}-{c}\right)^{T}{W}\left({x}_{i}-{c}\right)}$ makes
the gradient calculation even more intractable for the geometric loss
function. 

\subsection{Algebraic Loss}

However, analytic solutions for a sphere estimation can be derived
for algebraic loss. It can also generalize to ellipsoid (i.e., an
algebraic loss can be solved analytically for the ellipsoid ${x}^{T}{W}{x}=r$)
\begin{align*}
	\frac{\partial\mathscr{L}({c},r)}{\partial r} & =\sum_{i=1}^{n}\frac{\partial}{\partial r}d^{2}({x}_{i},S({c},r))\text{, algebraically}\\
	& =\sum_{i=1}^{n}\frac{\partial}{\partial r}\left(\left({x}_{i}-{c}\right)^{T}\left({x}_{i}-{c}\right)-r^{2}\right)^{2}\\
	& =\sum_{i=1}^{n}2\left(\left({x}_{i}-{c}\right)^{T}\left({x}_{i}-{c}\right)-r^{2}\right)\cdot(-2r)
\end{align*}
which is a cubic polynomial. $\frac{\partial\mathscr{L}({c},r)}{\partial r}=0$
is analytically solvable in $r$, via \href{https://en.wikipedia.org/wiki/Cubic_equation}{Cardano-Viete's formula}:
\begin{align*}
	0 & =\sum_{i=1}^{n}-2\left(\left({x}_{i}-{c}\right)^{T}\left({x}_{i}-{c}\right)-r^{2}\right)\cdot2r\\
	0 & =\sum_{i=1}^{n}\left(\left({x}_{i}^{T}{x}_{i}-2{c}^{T}{x}_{i}+{c}^{T}{c}\right)-r^{2}\right)\cdot r\\
	0 & =\sum_{i=1}^{n}\left(\left({x}_{i}^{T}{x}_{i}-2{c}^{T}{x}_{i}+{c}^{T}{c}\right)r-r^{3}\right)\\
	0 & =-n\cdot r^{3}+\left[\sum_{i=1}^{n}\left({x}_{i}^{T}{x}_{i}-2{c}^{T}{x}_{i}+{c}^{T}{c}\right)\right]\cdot r.
\end{align*}
Write it into $x^{3}+px+q=0$ form: 
\begin{align*}
	r^{3}+\left[-\frac{1}{n}\sum_{i=1}^{n}\left({x}_{i}^{T}{x}_{i}-2{c}^{T}{x}_{i}+{c}^{T}{c}\right)\right]\cdot r & +0=0\\
	p=-\frac{1}{n}\sum_{i=1}^{n}\left({x}_{i}^{T}{x}_{i}-2{c}^{T}{x}_{i}+{c}^{T}{c}\right) & ,q=0
\end{align*}
The determinant ${\displaystyle 4p^{3}+27q^{2}<0}$ obviously, the
solution is 
\begin{align*}
	\hat{r}_{k} & =2\sqrt{-\frac{p}{3}}\cdot\cos\left[\frac{1}{3}\arccos\left(\frac{3q}{2p}\sqrt{\frac{-3}{p}}\right)-\frac{2\pi k}{3}\right]\text{for}~k=0,1,2.\\
	& =2\sqrt{\frac{1}{3n}\sum_{i=1}^{n}\left({x}_{i}^{T}{x}_{i}-2{c}^{T}{x}_{i}+{c}^{T}{c}\right)}\cdot\cos\left[\frac{1}{3}\cdot\frac{\pi}{2}-\frac{2\pi k}{3}\right]
\end{align*}
For the gradient with respect to the center ${c}$,
\begin{align*}
	\frac{\partial\mathscr{L}({c},\hat{r})}{\partial{c}} & =\sum_{i=1}^{n}\frac{\partial}{\partial{c}}d^{2}({x}_{i},S({c},\hat{r}))\text{, algebraically}\\
	& =\sum_{i=1}^{n}\frac{\partial}{\partial{c}}\left(\left({x}_{i}-{c}\right)^{T}\left({x}_{i}-{c}\right)-\hat{r}^{2}\right)^{2}\\
	& =\sum_{i=1}^{n}2\left(\frac{\partial}{\partial{c}}\left[\left({x}_{i}-{c}\right)^{T}\left({x}_{i}-{c}\right)-\frac{1}{n}\sum_{j=1}^{n}\left({x}_{j}-{c}\right)^{T}\left({x}_{j}-{c}\right)\right]\right)\\
	& =\sum_{i=1}^{n}2\left(\frac{\partial}{\partial{c}}\left[\left({x}_{i}^{T}{x}_{i}-2{c}^{T}{x}_{i}+{c}^{T}{c}\right)-\frac{1}{n}\sum_{j=1}^{n}\left({x}_{j}^{T}{x}_{j}-2{c}^{T}{x}_{j}+{c}^{T}{c}\right)\right]\right),
\end{align*}
and the equation $\frac{\partial\mathscr{L}({c},\hat{r})}{\partial{c}}=0$
solves 
\begin{align*}
	\hat{{c}} & =\frac{1}{2}\left(\sum_{i=1}^{n}({x}_{i}-\frac{1}{n}\sum_{j=1}^{n}{x}_{j})^{T}({x}_{i}-\frac{1}{n}\sum_{j=1}^{n}{x}_{j})\right)^{-1}\sum_{i=1}^{n}\left({x}_{i}^{T}{x}_{i}-\frac{1}{n}\sum_{j=1}^{n}{x}_{j}^{T}{x}_{j}\right)({x}_{i}-\frac{1}{n}\sum_{j=1}^{n}{x}_{j}).
\end{align*}
Therefore, an algebraic loss would provide us a closed form solution
to the estimate of both center ${c}$ and radius $r$.%, once $V$ is determined. \hrluonote{serious check on the usage of V}
\subsection{SPCA and SRCA Solution}

Following the thought of the simultaneous estimation of ${c},r$ and
the dimension of the sphere (or equivalently, the linear subspace
${\bf V}\in \RR^{d\times (d'+1)}$ where $S$ lives in), we can instead consider the following geometric
loss in one step 
\begin{align*}
	\mathscr{L}({ V},{c},r) & =\sum_{i=1}^{n}d^{2}({x}_{i},S_{{ V}}({c},r))\\
	& =\sum_{i=1}^{n}d^{2}({x}_{i},{c}+V)+\sum_{i=1}^{n}d^{2}(Pr_{{  V}}({x}_{i}),S_{{ V}}({c},r))\\
	& =\sum_{i=1}^{n}\|{x}_{i}-{c}-{V}{V}^{T}({x}_{i}-{c})\|^{2}+\sum_{i=1}^{n}(\|Pr_{c+{ V}}({x}_{i})-{c}\|-r)^{2}\\
	& =\sum_{i=1}^{n}\|{x}_{i}-{c}-{V}{V}^{T}({x}_{i}-{c})\|^{2}+\sum_{i=1}^{n}(\|{c}+{V}{V}^{T}({x}_{i}-{c})-{c}\|-r)^{2}\\
	& =\sum_{i=1}^{n}\|{x}_{i}-{c}-{V}{V}^{T}({x}_{i}-{c})\|^{2}+\sum_{i=1}^{n}(\|{V}{V}^{T}({x}_{i}-{c})\|-r)^{2}
\end{align*}
The second identity comes from the Pythagorean theorem and $Pr_{c+{ V}}(x_{i})$
is the linear projection of $x_{i}$ to the affine subspace $c+{ V}$.

The first sum corresponds to PCA loss function and the second term
is the loss of SRCA if ${V}={I}$. For the SRCA and the SPCA, we minimize
the first sum so ${V}$ is the top eigenvectors of sample covariance
matrices and then plug this ${V}$ to the second sum, and change
the geometric sum to the algebraic loss function, since only the latter
loss allows a closed form analytic solution. This minimizer from a
two-step procedure obtained by SPCA is not necessarily the same
as the true minimizer of the above geometric loss $\mathscr{L}({ V},{c},r)$.
However, these two minimizers coincide when all ${x}_{i}$ are
from a sphere, otherwise SPCA solution is sub-optimal \citep{li_efficient_2022}.
We adopt the two-step SPCA algorithm only because we cannot derive
a closed form minimizer for $L=\mathscr{L}({V},{c},r)$. Moreover, this loss
function is difficult to generalize to the ellipsoid situation.
\section{\label{sec:SPCA-Asymptotics}Related Proofs}

\subsection{\label{proof Lip}Proof of Theorem \ref{thm:Lip}}
For each fixed $\|{v}\|_{l_{0}}=d'+1$, it suffices
to optimize the following sub-problem of (\ref{eq:SPCA_standard-1}):

\begin{align}
	& \min_{{c}\in\mathbb{R}^{d},r\in\mathbb{R}^{+}}\sum{}_{i=1}^{n}\left(({x}_{i}-{c})^{T} {W}({x}_i-{c})+r^{2}-2r\sqrt{\left({x}_{i}-{c}\right)^{T}\sqrt{{W}}^{T}{v}^{T}{I}{v}\sqrt{{W}}\left({x}_{i}-{c}\right)}\right) \\ =&\min_{{c}\in\mathbb{R}^{d},r\in\mathbb{R}^{+}}\mathscr{L}_{{v}}({c},r;{x}_{1},{x}_{2},\cdots,{x}_{n}),\nonumber & \\
	=&\min_{{c}\in\mathbb{R}^{d},r\in\mathbb{R}^{+}}\sum{}_{i=1}^{n}\mathscr{L}_{{v}}({c},r;{x}_{i}), & \label{eq:sub-problem}
\end{align}
which has gradients with respect to ${c}$ and $r$ as 

\begin{align*}
	& \frac{\partial \mathscr{L}_{{v}}}{\partial{c}}  =\sum{}_{i=1}^{n}\frac{\partial \mathscr{L}_{{v}}}{\partial{c}}({c},r;{x}_{i})\\
	& =\sum{}_{i=1}^{n}\left(-2({x}_{i}-{c})^{T}{W}-2r\cdot\frac{1}{2}\left[\left({x}_{i}-{c}\right)^{T}\sqrt{{W}}^{T}{v}^{T}{I}{v}\sqrt{{W}}\left({x}_{i}-{c}\right)\right]^{-\frac{1}{2}}\right. \\
	& \left.\left[-2\left({x}_{i}-{c}\right)^{T}\sqrt{{W}}^{T}{v}^{T}{I}{v}\sqrt{{W}}\right]\right)\\
	& =\sum{}_{i=1}^{n}-2({x}_{i}-{c})^{T}\left({W}+r\left[\left({x}_{i}-{c}\right)^{T}\sqrt{{W}}^{T}{v}^{T}{I}{v}\sqrt{{W}}\left({x}_{i}-{c}\right)\right]^{-\frac{1}{2}}\left[\sqrt{{W}}^{T}{v}^{T}{I}{v}\sqrt{{W}}\right]\right),
\end{align*}
%}
and, 

\begin{align*}
	\frac{\partial \mathscr{L}_{{v}}}{\partial r} & =\sum{}_{i=1}^{n}\frac{\partial \mathscr{L}_{{v}}}{\partial r}({c},r;{x}_{i})
	=\sum{}_{i=1}^{n}\left(2r-2\left[\left({x}_{i}-{c}\right)^{T}\sqrt{{W}}^{T}{v}^{T}{I}{v}\sqrt{{W}}\left({x}_{i}-{c}\right)\right]^{\frac{1}{2}}\right).
\end{align*}

Therefore, we can assume that the mild assumptions $\|{x}_{i}-{c}\|\leq R_{1},r\leq R_{2}$
and $|\lambda_{\max}({W})|\leq R_{3}$. We can compute the bounds of these gradients, using Cauchy-Schwartz inequality in the first inequality: 
%{\scriptsize 
\begin{align*}
	& \|\nabla_{({c},r)}\mathscr{L}_{{v}}({c},r)\|  =\left\|\frac{\partial \mathscr{L}_{{v}}}{\partial{c}}({c},r)\right\|+\left\|\frac{\partial \mathscr{L}_{{v}}}{\partial r}({c},r)\right\|\\
	& \leq4\sum_{i=1}^{n}({x}_{i}-{c})^{T} {W}^T{W}({x}_i-{c}) \\
	& \times\sum_{i=1}^{n}\left\Vert \left({W}+r\left[\left({x}_{i}-{c}\right)^{T}\sqrt{{W}}^{T}{v}^{T}{I}{v}\sqrt{{W}}\left({x}_{i}-{c}\right)\right]^{-\frac{1}{2}}\left[\sqrt{{W}}^{T}{v}^{T}{I}{v}\sqrt{{W}}\right]\right)\right\Vert^2 \\
	& +\sum{}_{i=1}^{n}\left(2r-2\left[\left({x}_{i}-{c}\right)^{T}\sqrt{{W}}^{T}{v}^{T}{I}{v}\sqrt{{W}}\left({x}_{i}-{c}\right)\right]^{\frac{1}{2}}\right).\\
	& \leq4\times 2nR^2_3 \times nR_{1}^{2}\times n\left(R_3+R_{2}\frac{\sqrt{R_{3}^{2}}}{\sqrt{R_{1}^{2}}}\right)+n\left(2R_{2}+\sqrt{R_{1}^{2}R_{3}^{2}}\right)\\
	& <\infty
\end{align*}
%}
For a finite $n$, we can conclude that $\mathscr{L}$ is Lipschitz with a finite
Lipschitz constant as bounded above. Then the gradient descent algorithm
would give us a solution to the sub-problem (\ref{eq:sub-problem})
with linear convergence from classical results \citep{boyd2004convex}. Since for fixed ${v}$, each sub-problem converges to the solution,
the exhaustive search on ${v}$ solves the original problem (\ref{eq:SPCA_standard-1}). In parallel to \citet{boyd2003subgradient}, we have proved the Theorem \ref{thm:Lip}.

\iffalse
\begin{itemize}
	\item \citet{huber2004robust} Robust Statistics, 2ed, 2004.
	We mainly use the materials from Chapters 3 and 6 in this book, but the essence of the
	proof is in \citet{huber1967behavior}.
	\item \citet{huber1967behavior} The behavior of maximum
	likelihood estimates under nonstandard conditions, 1967. (Excerpt
	from the 5th Berkeley Symposium)
	\item \citet{doob1953stochastic} Stochastic Processes,
	1953. (Wiley reproduced version, 1990)
\end{itemize}
\fi
\subsection{\label{sec:Cor_convergence}Proof of Theorem   \ref{thm:consistency_clean}}
It is clear that $\mathscr{L}({ c_0},r_0,\mathcal{I}_0)=0$ and $\widehat{\mathcal{I}}_k,\widehat{c}_k,\widehat{r}_k\to \arg\min \mathscr{L}$ by Theorem \ref{thm:Lip}, it suffices to show $({c_0},r_0,\mathcal{I}_0)$ is the unique zero of $L$. Recall that $\mathscr{L}({ c},r,\mathcal{I})=0$ if and only if all all ${x}_i$'s are exactly on sphere $S({c},r,\mathcal{I})$, and that $d'+2$ points uniquely determine a $d'$ dimensional sphere, then the uniqueness follows from the assumption $n>d'+1$.

\subsection{\label{Gamma_noise proof}Proof of Theorem \ref{thm:consistency_noise}}
We consider the closed set $\Theta_{1}$ on the parameter space defined
by $\|x_{i}-c\|\leq R_{1},~\forall i=1\cdots,n,~r\leq R_{2}$ and
$|\lambda_{\max}(W)|\leq R_{3}$ as we did in the proof of Theorem \ref{Theorem A} and \ref{Theorem B}. 

Again, let us assume $\mathcal{I}$ to be fixed index set and the 
\[
f_{\infty}(c,r)=\lim_{n\to\infty}\frac{1}{n}\sum_{i=1}^{n}\left((y_{i}-c)^{T}W(y_{i}-c)-r-2r\sqrt{(y_{i}-c)^{T}\sqrt{W}^{T}I_{\mathcal{I}}\sqrt{W}(y_{i}-c)}\right)^{2}
\]
 be the limiting form of our geometric loss function, 
\[
\mathcal{L}(c,r,\mathcal{I}\mid\mathcal{Y})=f_{j}(c,r)=\frac{1}{j}\sum_{i=1}^{j}\left((y_{i}-c)^{T}W(y_{i}-c)-r-2r\sqrt{(y_{i}-c)^{T}\sqrt{W}^{T}I_{\mathcal{I}}\sqrt{W}(y_{i}-c)}\right)^{2}
\]
 with the dataset $\mathcal{Y}=\{y_{1},\cdots,y_{j}\}$ and 
\[
\mathcal{L}(c,r,\mathcal{I}\mid\mathcal{X})=g_{j}(c,r)=\frac{1}{j}\sum_{i=1}^{j}\left((x_{i}-c)^{T}W(x_{i}-c)-r-2r\sqrt{(x_{i}-c)^{T}\sqrt{W}^{T}I_{\mathcal{I}}\sqrt{W}(x_{i}-c)}\right)^{2}
\]
 with the dataset $\mathcal{X}=\{x_{1},\cdots,x_{j}\}$. Recall that
$x_{i}=y_{i}+\epsilon_{i}$ and $y_{i}\in S_{W}(c_{0},r_{0})$ lying
on an ellipsoid with center $c_{0}$, radius $r_{0}$ and known covariance
$W$. 

Therefore, $\arg\min f_{\infty=}\arg\min f_{j}=(c_{0},r_{0})$ since if we plug in
$c_{0}$ and $r_{0}$ the $f_{\infty}(c_{0},r_{0})=f_{j}(c_{0},r_{0})=0$. By the definition of $f_{\infty}$, $f_{j}$ converges to $f_\infty$ point-wise.
In addition, by the compact assumptions, the convergence is also uniform, that is, $\sup_{\theta\in\Theta_1}|f_j(\theta)-f_\infty(\theta)|\to 0$. 

The rest of our roadmap of proof is as follows. According to the Remark
1.10 of \citet{braides2002gamma}: if a sequence of functions $g_{j}$ point-wisely
converges to its limit $f_{\infty}$ uniformly. and $f_{\infty}$
is lower semi-continous, then the same sequence of functions also
converges in a $\Gamma$-convergence sense, and its $\Gamma$-limit
is identical to its point-wise limit $f_{\infty}=\lim_{j\to\infty}f_{j}$
. Furthermore, as we showed above, the following minimizer exists $$\theta_{*}\coloneqq\arg\min_{\theta\in\Theta_{1}}f_{\infty}(\theta).$$
Then since the specific form of our geometric loss function \eqref{eq:basic opt with c and r} is coercive,
by Theorem 1.21 and Remark 1.22 in \citet{braides2002gamma}, the minimizer sequence
$\{\theta_{j}\}=\{\arg\min_{\theta\in\Theta_{1}}f_{j}(\theta)\}$
converges to a minimum point $\theta_{*}$ of $f_{\infty}$. Note
that our assumption (A1) stating that each of $\theta_{j}\coloneqq\arg\min_{\theta\in\Theta_{1}}g_{j}(\theta) $ 
exist is essential here. Otherwise the sequence will not exist.

It's it clear that $f_{j}$ uniformly converges to $f_{\infty}$ on compact set $\Theta_{1}$.
We focus on proving $g_{j}$ also converges to $f_{j}$ uniformly,
then through a middle-man argument, $\lim_{j\rightarrow\infty}g_{j}=f_{\infty}$
holds. The difference between two sequences ${f_j}$ and ${g_j}$ can be bounded as below: 
\begin{align}
 & \left|g_{j}(c,r)-f_{j}(c,r)\right|\\
 & =\left|\frac{1}{j}\sum_{i=1}^{j}(y_{i}-c)^{T}W(y_{i}-c)-(x_{i}-c)^{T}W(x_{i}-c)+\right.\nonumber\\
 & \left.2r\sqrt{(x_{i}-c)^{T}\sqrt{W}^{T}I_{\mathcal{I}}\sqrt{W}(x_{i}-c)}-2r\sqrt{(y_{i}-c)^{T}\sqrt{W}^{T}I_{\mathcal{I}}\sqrt{W}(y_{i}-c)}\right|\\
 & \leq\left(4R_{3}^{2}+R_{1}R_{3}\right)\left|\frac{1}{j}\sum_{i=1}^{j}(\|x_{i}-c\|+\|y_{i}-c\|-2r)(\|x_{i}-c\|-\|y_{i}-c\|)\right|\\
 & =\left(4R_{3}^{2}+R_{1}R_{3}\right)\left|\frac{1}{j}\sum_{i=1}^{j}(\|x_{i}-c\|+\|y_{i}-c\|-2r)(\|y_{i}-c+\epsilon_{i}\|-\|y_{i}-c\|)\right|\nonumber\\
 & \leq\left(4R_{3}^{2}+R_{1}R_{3}\right)\frac{1}{j}\sum_{i=1}^{j}|(\|x_{i}-c\|+\|y_{i}-c\|-2r)|(\|\epsilon_{i}\|+2\|\epsilon_{i}\|\|y_{i}-c\|)\\
 & =\left(4R_{3}^{2}+R_{1}R_{3}\right)\frac{1}{j}\sum_{i=1}^{j}|(\|x_{i}-c\|+\|y_{i}-c\|-2r)|(\|\epsilon_{i}\|+2\|y_{i}-c\|)\cdot\|\epsilon_{i}\|\nonumber\\
 & \leq\left(4R_{3}^{2}+R_{1}R_{3}\right)\cdot\left(2R_{1}+2R_{2}\right)\cdot\left(1+2R_{1}\right)\cdot\frac{1}{j}\sum_{i=1}^{j}\|\epsilon_{j}\|,\label{eq:keyeqA3}
\end{align}
where we need the assumption (A2) stating that $x_{i},y_{i}$ and
$c,r$ are all in the closed bounded (hence compact) set $B\times\Theta_{1}$.
We want to show that as $j\rightarrow\infty$, 
\[
\sup_{\theta=(c,r)\in\Theta_{1}}\|g_{j}-f_{j}\|\to0,
\]
to ensure uniform convergence. But this follows from \eqref{eq:keyeqA3} and our assumption (A3) stating that $\lim_{n\to\infty}\frac{1}{n}\sum_{i=1}^{n}\|\epsilon_{i}\|=0$.
Such an assumption is common, see, \citet{maggioni2016multiscale,fefferman2018fitting,aamari2019nonasymptotic}
for instances. In fact, the assumption is even weaker than those in
the above references, for example, in \citet{aamari2019nonasymptotic}
the amplitude of the noise is assume to be $\|\epsilon\|\sim n^{-\frac{\alpha}{d}}$
for $\alpha>1$. In contrast, we only require $\|\epsilon\|\to0$,
so $\|\epsilon\|\sim n^{-\alpha}$ for any $\alpha>0$ or even $\|\epsilon\|\sim\frac{1}{\log n}$
is good enough.

\subsection{\label{Con proof}Proof of Theorem \ref{Theorem A}}
References we mainly need for our proof below are the formulation in \citet{huber2004robust,huber1967behavior} and the technical separation lemma in \citet{doob1953stochastic}.

We fix the index set $\mathcal{I}$ in the following discussions,
and we assume that the parameters to be estimated can be written as
a vector $\theta=({c},r)\in\Theta\coloneqq[-C,C]^{d}\times[R_0,R]\subset\mathbb{R}^{d}\times\mathbb{R}^{+}$,
which lies in a (locally) compact space with a countable base $\Theta'=\left\{ [-C,C]^{d}\cap\mathbb{Q}^{d}\right\} \times\left\{ [R_0,R]\cap\mathbb{Q}\right\} $,
the inclusion of $r=R_0$ is needed below for compactness. We denote
that estimate for $\theta$ based on $n$ samples (by minimization
of the $\mathscr{L}$) by $T_{n}=T_{n}(\mathcal{X})$. 

The real-valued $\rho$ function, based on the samples ${x}_{1},\cdots,{x}_{n}\in\mathbb{X}=\mathbb{R}^{d}$
drawn from the common distribution $P$ defined on the probability
space $(\mathbb{X},\mathcal{A},\nu)$ with Borel algebra $\mathcal{A}$
and Lebesgue measure $\nu$, is 
\[
\rho({x};\theta)=\left(({x}-{c})^{T} {W}({x}-{c})+r^{2}-2r\sqrt{\left({x}-{c}\right)^{T}\sqrt{{W}}^{T}{I}_{\mathcal{I}}\sqrt{{W}}\left({x}-{c}\right)}\right)
\]
and the $\psi({x};\theta)=\frac{\partial}{\partial\theta}\rho({x};\theta)$
is again differentiable. We show below that the assumptions in \citet{huber1967behavior}
are satisfied, we define our estimator $T_{n}$ for parameter $\theta=({c},r)$
such that 
\begin{align*}
	\frac{1}{n}\sum_{i=1}^{n}\rho({x}_{i};T_{n})-\inf_{\theta\in\Theta}\frac{1}{n}\sum_{i=1}^{n}\rho({x}_{i};\theta) & \rightarrow0,\text{a.s. }P & \text{ when }n \rightarrow\infty,
\end{align*}
corresponding to case A in \citet{huber1967behavior}.
Since $\rho$ is differentiable in both ${x},\theta$, this minimizer
could also be expressed in form of $T_{n}$ satisfying 
\begin{align*}
	\frac{1}{\sqrt{n}}\sum_{i=1}^{n}\psi({x}_{i};T_{n}) & \rightarrow0,\text{a.s. }P & \text{ when }n \rightarrow\infty,
\end{align*}

\begin{itemize}%[leftmargin=*]
	\item (A-1) For $\Theta=\mathbb{R}^{d}\times\mathbb{R}^{+}$, there exists
	a countable basis $\Theta'=\left\{ [-C,C]^{d}\cap\mathbb{Q}^{d}\right\} \times\left\{ [R_0,R]\cap\mathbb{Q}\right\} $
	such that for every open set $U\subset\Theta$ and every closed interval
	$A\subset\mathbb{R}$, two sets 
	\begin{align*}
		\{{x} & \mid\rho({x};\theta)\in A \in \mathcal{A},\forall\theta\in U\}\\
		\{{x} & \mid\rho({x};\theta)\in A \in \mathcal{A},\forall\theta\in U\cap\Theta'\}
	\end{align*}
	would only differ on the set of zero probability measure $P$.\textbf{
	}Since the measure $P$ is fixed, by Lemma 2.1 on page 56 of \citet{doob1953stochastic},
	for each $\theta\in\Theta$ we can find $\theta'\in\Theta'$ such
	that 
	\[
	P\left\{ \omega\in\mathbb{X}\mid\rho({x}(\omega);\theta)\neq\rho({x}(\omega);\theta'),{x}(\omega)\sim P\right\} =0.
	\]
	Therefore, denote the map $\tau_{P}:\theta\mapsto\theta'$ we can
	redefine our $\rho$ by $\tilde{\rho}\coloneqq\rho\circ\tau_{P}$
	so that it only differs from $\rho$ on a zero measure set of the
	fixed $P$. Note that the mapping $\tau_{P}$ depends on the measure
	$P$ and we assume $P$ is fixed throughout our discussion. This
	ensures the measurability of $\inf_{\theta'\in U}\tilde{\rho}({x};\theta')$
	and the measurability of its limit when an (open) neighbor hood $U$
	of $\theta$ shrinks to one-point set $\{\theta\}$. For ease of notation,
	we still use $\rho$ below as assume (A-1) holds. 
	\item (A-2) The function $\rho$ is continuous and differentiable, therefore
	clearly lower semi-continuous in $\theta=({c},r)$. And this ensures
	that $\inf_{\theta'\in U}\rho({x};\theta')\rightarrow\rho({x};\theta)$. 
	\item (A-3) There exists a measurable function $a({x})$ such that 
	\begin{align*}
		\mathbb{E}_{P}\left(\rho({x};\theta)-a({x})\right)^{-} & <\infty\\
		\mathbb{E}_{P}\left(\rho({x};\theta)-a({x})\right)^{+} & <\infty
	\end{align*}
	and hence $\gamma(\theta)=\mathbb{E}\left(\rho({x},\theta)-a({x})\right)$
	is well-defined for all $\theta\in \Theta$. For our purpose, we choose
	$\theta_{1}=({c}_{1},r_{1})$ for some $\|{c}_{1}\|<\infty$
	and $r_{1}<\infty$. We define a function on $\mathbb{X}$
	\begin{align*}
		a({x})  =a_{\theta_{1}}({x}) & =\left(({x}-{c}_1)^{T} {W}({x}-{c}_1)+r_{1}^{2}-2\cdot r_{1}\sqrt{\left({x}-{c}_{1}\right)^{T}\sqrt{{W}}^{T}{I}_{\mathcal{I}}\sqrt{{W}}\left({x}-{c}_{1}\right)}\right)\\
		\rho({x};\theta)-a({x}) & =\left(({x}-{c})^{T} {W}({x}-{c})-({x}-{c}_1)^{T} {W}({x}-{c}_1)\right)+\left(r^{2}-r_{1}^{2}\right)\\
		& -2r\sqrt{\left({x}-{c}\right)^{T}\sqrt{{W}}^{T}{I}_{\mathcal{I}}\sqrt{{W}}\left({x}-{c}\right)}\\
		& +2r_{1}\sqrt{\left({x}-{c}_{1}\right)^{T}\sqrt{{W}}^{T}{I}_{\mathcal{I}}\sqrt{{W}}\left({x}-{c}_{1}\right)}\\
		& \leq|\lambda_{\max}({W})|\left(\|{x}-{c}\|^{2}-\|{x}-{c}_{1}\|^{2}\right)+\left(r^{2}-r_{1}^{2}\right)\\
		& +4\max(r,r_{1})\cdot\max(\|{c}\|,\|{c}_{1}\|)\cdot\left|\lambda_{\max}({W})\right|\cdot\|{x}\|
	\end{align*}
	If we take $\mathbb{E}_{P}$ on both sides of inequality above and
	with the assumption that $|\lambda_{\max}({W})|<R_3$,
	then the mild assumption that $P$ has finite second moments (hence
	finite first moment) ensures the finiteness. It is not hard
	to see that the choice of $\theta_{1}$ is not essential in verifying
	this assumption. For simplicity, we assume $\theta_{1}=({c}_{1},r_{1})=({0},1)$
	hereafter. 
	\begin{align*}
		a({x}) & =\left(|\lambda_{\max}({W})|\|{x}\|^{2}+1-2\sqrt{{x}^{T}\sqrt{{W}}^{T}{I}_{\mathcal{I}}\sqrt{{W}}{x}}\right)
	\end{align*}
	\label{true parameter}
	\item (A-4) There is a $\theta_{0}\in\Theta$ such that $\gamma(\theta)>\gamma(\theta_{0})$
	for all $\theta\neq\theta_{0}$. To see this, we use the Fubini theorem
	to take differentiation inside the $\mathbb{E}_{P}$ (note that this
	is taken with respect to ${x}$) to conclude unique minima of $\gamma(\theta)$
	(notice that we assume $\mathcal{I}$ fixed and therefore the index
	vector ${v}$ is a fixed constant vector) 
	%{\scriptsize 
	\begin{align*}
		& \gamma(\theta)=\mathbb{E}_{P}\rho({x};\theta)-a({x})\\
		& =\mathbb{E}_{P}\left(({x}-{c})^{T} {W}({x}-{c})+r^{2}-2r\sqrt{\left({x}-{c}\right)^{T}\sqrt{{W}}^{T}{I}_{\mathcal{I}}\sqrt{{W}}\left({x}-{c}\right)}\right)-a({x})\\
		& \frac{\partial}{\partial\theta}\gamma(\theta)=\mathbb{E}_{P}\left(\begin{array}{c}
			\frac{\partial}{\partial{c}}\rho({x};\theta)\\
			\frac{\partial}{\partial r}\rho({x};\theta)
		\end{array}\right)\\
		& =\left(\begin{array}{c}
			-\mathbb{E}_{P}2({x}-{c})^{T}\left({W}+r\left[\left({x}-{c}\right)^{T}\sqrt{{W}}^{T}{v}^{T}{I}_{p}{v}\sqrt{{W}}\left({x}-{c}\right)\right]^{-\frac{1}{2}}\left[\sqrt{{W}}^{T}{v}^{T}{I}_{p}{v}\sqrt{{W}}\right]\right)\\
			\mathbb{E}_{P}2r-2\left[\left({x}-{c}\right)^{T}\sqrt{{W}}^{T}{v}^{T}{I}_{p}{v}\sqrt{{W}}\left({x}-{c}\right)\right]^{\frac{1}{2}}
		\end{array}\right)\\
		& ={0}
	\end{align*}
	%}
	By letting $\frac{\partial}{\partial\theta}\gamma(\theta)=0$ and
	for ${x}\sim P$, we derive from the second equation that 
	\[
	r_{0}({c})=\mathbb{E}_{P}\left[\left({x}-{c}\right)^{T}\sqrt{{W}}^{T}{v}^{T}{I}_{p}{v}\sqrt{{W}}\left({x}-{c}\right)\right]^{\frac{1}{2}}\in[0,\min(R,2C\sqrt{|\lambda_{\max}({W})|})],
	\]
	and from the first equation 
	%{\scriptsize 
	\begin{align*}
		\mathbb{E}_{P}2\left({x}-{c}\right)^{T}\left({W}+r_{0}({c})\left[\left({x}-{c}\right)^{T}\sqrt{{W}}^{T}{v}^{T}{I}_{p}{v}\sqrt{{W}}\left({x}-{c}\right)\right]^{-\frac{1}{2}}\left[\sqrt{{W}}^{T}{v}^{T}{I}_{p}{v}\sqrt{{W}}\right]\right) & ={0}
	\end{align*}
	%}
	Consider the following function
	%{\scriptsize 
	\begin{align}
		\ensuremath{{F}({c})}\ensuremath{\coloneqq}\ensuremath{\mathbb{E}_{P}}2\ensuremath{\left({x}-{c}\right)^{T}\left({W}+r_{0}({c})\left[\left({x}-{c}\right)^{T}\sqrt{{W}}^{T}{v}^{T}{I}_{p}{v}\sqrt{{W}}\left({x}-{c}\right)\right]^{-\frac{1}{2}}\right.\nonumber\\
			\left.\left[\sqrt{{W}}^{T}{v}^{T}{I}_{p}{v}\sqrt{{W}}\right]\right)}
	\end{align}
	% }
	as a function of ${c}$ and the above equation becomes ${F}({c})={0}$.
	Taking a sandwiching-style argument, we first note that the second term in the second bracket is always non-negative, then we construct uniform bounding functions: 
	%{\scriptsize 
	\begin{align*}
		& {F}_{1}({c})  \coloneqq\mathbb{E}_{P}2\left({x}-{c}\right)^{T}{W}\\& \asymp\mathbb{E}_{P}\left({x}-{c}\right)^{T}{W},\\
		& {F}_{2}({c})  \coloneqq\mathbb{E}_{P}2\left({x}-{c}\right)^{T}\left({W}+\min(R,2C\sqrt{|\lambda_{\max}({W})|})\cdot\right.\\
		& \left.|\lambda_{\max}({W})|\left[\left({x}-{c}\right)^{T}\sqrt{{W}}^{T}{v}^{T}{I}_{p}{v}\sqrt{{W}}\left({x}-{c}\right)\right]^{-\frac{1}{2}}\right){W}\\& \asymp\mathbb{E}_{P}\left({x}-{c}\right)^{T}\left(1+\frac{K(R,C,{v},|\lambda_{\max}({W})|)}{\|{x}-{c}\|_{2}}\right){W},
	\end{align*}
	%}
	(where $K(R,C,{v},|\lambda_{\max}({W})|)$ is a non-negative constant) such
	that the following bound ${F}_{1}({c})\leq{F}({c})\leq{F}_{2}({c})$
	holds (for each component of the vector-valued ${F}_1,{F}_2$) uniformly in ${c}$. However, it is clear that there exists
	${c}_{1}^{+},{c}_{2}^{-}\in[-C,C]^{d}\subset\mathbb{R}^{d}$
	\begin{align*}
		{F}({c}_{1}^{+})\geq{F}_{1}({c}_{1}^{+})>0,\\
		{F}({c}_{2}^{-})\leq{F}_{2}({c}_{2}^{-})<0.
	\end{align*}
	Note that ${F}$ is continuous in ${c}$ (we can take derivative
	under $\mathbb{E}_{P}$ since $P$ is assumed to possess finite second
	moment) and $[-C,C]^{d}$ is connected, we apply the multivariate
	intermediate value theorem to assert the existence of \textbf{a} solution
	${c}_{0}$ for ${F}({c})={0}$. Therefore, we can keep
	this solution ${c}_{0}$, which we know its existence but do not
	know its expression. Back substitution of this solution of ${c}_{0}$
	into the expression of $r_{0}$ yields 
	\begin{align*}
		r_{0}= & \mathbb{E}_{P}\left[\left({x}-{c}_{0}\right)^{T}\sqrt{{W}}^{T}{v}^{T}{I}_{p}{v}\sqrt{{W}}\left({x}-{c}_{0}\right)\right]^{\frac{1}{2}},
	\end{align*}
	where $\theta_{0}=({c}_{0},r_{0})$ is well-defined for $P$ with
	finite second moment. This verifies the assumption (A-4). 
	\item (A-5) With the notations in (A-3), since
	$\Theta\coloneqq[-C,C]^{d}\times[R_0,R]\subset\mathbb{R}^{d}\times\mathbb{R}^{+}$
	is a compact space, it suffices to verify only (i) of (A-5). There
	is a continuous function $b(\theta)>0$ such that 
	\begin{align*}
		b(\theta) & =\left({c}^{T}{W}{c}+r^{2}-2r\sqrt{{c}^{T}\sqrt{{W}}^{T}{I}_{\mathcal{I}}\sqrt{{W}}{c}}\right)+1\in [1,1+C^2|\lambda_{\max}({W})|+R^2]
	\end{align*}
	For a fixed ${x}\in\mathbb{X}$, the function $g_{{x}}(\theta)=2r\sqrt{\left({x}-{c}\right)^{T}\sqrt{{W}}^{T}{I}_{\mathcal{I}}\sqrt{{W}}\left({x}-{c}\right)}$
	has gradient 
	\begin{align*}
		\frac{\partial}{\partial\theta}g_{{x}}({c},r) & =\left(\begin{array}{c}
			\frac{\partial}{\partial{c}}g_{{x}}({c},r)\\
			\frac{\partial}{\partial r}g_{{x}}({c},r)
		\end{array}\right)\\
		& =\left(\begin{array}{c}
			r\left[\left({x}-{c}\right)^{T}\sqrt{{W}}^{T}{I}_{\mathcal{I}}\sqrt{{W}}\left({x}-{c}\right)\right]^{-\frac{1}{2}}\cdot2\left({x}-{c}\right)^{T}\sqrt{{W}}^{T}{I}_{\mathcal{I}}\sqrt{{W}}\\
			2\sqrt{\left({x}-{c}\right)^{T}\sqrt{{W}}^{T}{I}_{\mathcal{I}}\sqrt{{W}}\left({x}-{c}\right)}
		\end{array}\right)
	\end{align*}
	which is bounded from above in matrix norm by $2R\sqrt{|\lambda_{\max}({W})|}\cdot4RC\sqrt{|\lambda_{\max}({W})|}\leq16\max(R^{2},1)\cdot C|\lambda_{\max}({W})|\eqqcolon L_{g}<\infty$.
	Therefore, the function $g_{{x}}({c},r)$ is a $L_{g}$-Lipschitz function.
	We have 
	\begin{align*}
		\inf_{\theta\in\Theta}\frac{\rho({x};\theta)-a({x})}{b(\theta)} & =\inf_{\theta\in\Theta}\left\{ \left(({x}-{c})^T{W}({x}-{c})-{x}^{T}{W}{x}\right)+\left(r^{2}-1\right)\right.\\
		& -2r\sqrt{\left({x}-{c}\right)^{T}\sqrt{{W}}^{T}{I}_{\mathcal{I}}\sqrt{{W}}\left({x}-{c}\right)}\left.+2\sqrt{{x}^{T}\sqrt{{W}}^{T}{I}_{\mathcal{I}}\sqrt{{W}}{x}}\right\} \\
		& \left({c}^T{W}{c}+r^{2}-2r\sqrt{{c}^{T}\sqrt{{W}}^{T}{I}_{\mathcal{I}}\sqrt{{W}}{c}}+1\right)^{-1}\\
		& \geq\inf_{\theta=({c},r)\in\Theta}\left\{ \left(({x}-{c})^T{W}({x}-{c})-{x}^{T}{W}{x}\right)+\left(r^{2}-1\right)\right.\\
		& -2r\sqrt{\left({x}-{c}\right)^{T}\sqrt{{W}}^{T}{I}_{\mathcal{I}}\sqrt{{W}}\left({x}-{c}\right)}\left.+2\sqrt{{x}^{T}\sqrt{{W}}^{T}{I}_{\mathcal{I}}\sqrt{{W}}{x}}\right\} \\
		& (1+C^2|\lambda_{\max}({W})|+R^2)^{-1}  \eqqcolon h({x})
	\end{align*}
	and $\frac{\rho({x},\theta)-a({x})}{b(\theta)}\geq h({x})$
	by the infimum in the definition while $h({x})$ is integrable
	with respect to $P$ due to the fact that $g_{{x}}({c},r)$
	is Lipschitz.
\end{itemize}
Now we verify all assumptions (A-1) to (A-5) in \citet{huber1967behavior},
Theorem 1 in the same paper ensures that Theorem A holds.
\iffalse
\textbf{Theorem A.} Suppose that the index vector $\mathcal{I}$ is
fixed and the parameter $\theta=({c},r)\in\Theta\coloneqq[-C,C]^{d}\times[0,R]\subset\mathbb{R}^{d}\times\mathbb{R}^{+}$
for some $C,R\in(0,\infty)$ and the samples ${x}_{1},\cdots,{x}_{n}\in\mathbb{X}=\mathbb{R}^{d}$
of size $n$ are drawn from the common distribution $P$. $P$ has
finite second moments on the probability space $(\mathbb{X},\mathcal{A},\nu)$
with Borel algebra $\mathcal{A}$ and Lebesgue measure $\nu$. Then
the estimator $T_{n}$ defined by 
\begin{align*}
	\frac{1}{n}\sum_{i=1}^{n}\rho({x}_{i};T_{n})-\inf_{\theta\in\Theta}\frac{1}{n}\sum_{i=1}^{n}\rho({x}_{i};\theta) & \rightarrow0,\text{a.s. }P\\
	n & \rightarrow\infty,
\end{align*}
would converge in probability and almost surely to $\theta_{0}$ w.r.t.
$P$. Particularly, $T_{n}$ can be realized as a solution to our
optimization above. $\square$
\fi
The mild assumptions that $\theta$ lies in a compact subspace of
$\mathbb{R}^{d}\times\mathbb{R}^{+}$ can be relaxed by verifying
a more stringent set of conditions (A-5) as pointed out by \citet{huber1967behavior}.
Since we actually verify assuming that the index set $\mathcal{I}$
is fixed, we need to point out that in the $l_{0}$ optimization for
each fixed $\mathcal{I}$ the consistency result holds. But for the
$l_{1}$ relaxed problem, we cannot guarantee consistency even with
stronger assumptions, only algorithmic convergence is guaranteed.

\subsection{\label{AsyNor proof}Proof of Theorem \ref{Theorem B}}
Now we take the second view that the estimator sequence $T_{n}$ for
parameter $\theta=({c},r)$ and assume a fixed index set $\mathcal{I}$
such that 
\begin{align*}
	\frac{1}{n}\sum_{i=1}^{n}\psi({x}_{i};T_{n}) & \rightarrow0,\text{a.s. }P\\
	n & \rightarrow\infty,
\end{align*}

\begin{itemize}%[leftmargin=*]
	\item (N-1) For each fixed $\theta\in\Theta$, $\psi({x};\theta)$ is
	$\mathcal{A}$-measurable and separable. Like the construction in
	(A-1), we can modify the $\psi$ into a separable version $\tilde{\psi}$
	if necessary and verify this assumption. Then following functions
	are well-defined (with finite second moment assumption on $P$ and the Fubini theorem)
	\begin{align*}
		\lambda(\theta)=\lambda({c},r) & \coloneqq\mathbb{E}_{P}\psi({x};\theta)\\
		& =\mathbb{E}_{P}\frac{\partial}{\partial\theta}\rho({x};\theta)\\
		& =\frac{\partial}{\partial\theta}\mathbb{E}_{P}\rho({x};\theta)\\
		%& =\frac{\partial}{\partial\theta}\gamma(\theta)\\
		u({x},\theta,D) & =\sup_{\|\tau-\theta\|\leq D}\left|\psi({x};\tau)-\psi({x};\theta)\right|.
	\end{align*}
	
	\item (N-2) The same $\theta_{0}$ as computed above would satisfy $\lambda(\theta_{0})=0$.
	\item (N-3) There are strictly positive numbers $\alpha,\beta,\gamma,\eta$
	such that 
	\begin{itemize}
		\item (i) $\left|\lambda(\theta)\right|\geq\alpha|\theta-\theta_{0}|$ for
		some $\alpha>0$ and $|\theta-\theta_{0}|\leq\eta$ is clear since
		$$\lambda(\theta)=\frac{\partial}{\partial\theta}\mathbb{E}_{P}\left(({x}-{c})^T{W}({x}-{c})+r^{2}-2r\sqrt{\left({x}-{c}\right)^{T}\sqrt{{W}}^{T}{I}_{\mathcal{I}}\sqrt{{W}}\left({x}-{c}\right)}\right)$$
		is quadratic in both ${c}$ and $r$, and it is bounded from below
		by linear part due to Taylor expansion at $\theta_{0}$.
		\item (ii) $\mathbb{E}_{P}u({x},\theta,D)=\mathbb{E}_{P}\sup_{\|\tau-\theta\|\leq D}\left|\psi({x};\tau)-\psi({x};\theta)\right|\leq\mathbb{E}_{P}\beta\|\tau-\theta\|$
		since $\frac{\partial}{\partial r}\psi({x};\theta)=2$ and 
		
		%\begin{adjustbox}{center}
		% {\scriptsize 
		\begin{align*}
			\frac{\partial}{\partial{c}}\psi({x};\theta) & =\mathbb{E}_{P}2{c}{}^{T}\left({W}+r\left[\left({x}-{c}\right)^{T}\sqrt{{W}}^{T}{v}^{T}{I}{v}\sqrt{{W}}\left({x}-{c}\right)\right]^{-\frac{1}{2}}\left[\sqrt{{W}}^{T}{v}^{T}{I}{v}\sqrt{{W}}\right]\right) \\
			& -\mathbb{E}_{P}2({x}-{c})^{T}(-\frac{1}{2}r\left[\left({x}-{c}\right)^{T}\sqrt{{W}}^{T}{v}^{T}{I}{v}\sqrt{{W}}\left({x}-{c}\right)\right]^{-\frac{3}{2}} \\
			& \cdot2\left({x}-{c}\right)^{T}\sqrt{{W}}^{T}{v}^{T}{I}{v}\sqrt{{W}}\cdot\left[\sqrt{{W}}^{T}{v}^{T}{I}{v}\sqrt{{W}}\right]) \\
			& \leq2CR(1+(4C^{2}|\lambda_{\max}({W})|{}^{-\frac{1}{2}})|\lambda_{\max}({W})| \\
			& +4C\left(R\cdot(4C^{2}|\lambda_{\max}({W})|)^{-\frac{3}{2}}\cdot2C|\lambda_{\max}({W})|^{2}\right) \\
			& \leq16CR(4C^{2}|\lambda_{\max}({W})|{}^{-\frac{1}{2}}\max(|\lambda_{\max}({W})|,1)^{4} + |\lambda_{\max}({W})|<\infty. \\
		\end{align*}
		%}
		%\end{adjustbox}
		
		And $\psi({x};\theta)$ is Lipschitz with coefficient $$\beta\coloneqq32CR(4C^{2}|\lambda_{\max}({W})|{}^{-\frac{1}{2}}\max(|\lambda_{\max}({W})|,1)^{4}+|\lambda_{\max}({W})|.$$ 
		\item (iii) $\mathbb{E}_{P}u({x},\theta,D)^{2}=\mathbb{E}_{P}\left(\sup_{\|\tau-\theta\|\leq D}\left|\psi({x};\tau)-\psi({x};\theta)\right|\right)^{2}\leq$\\
		$\max\left\{ \mathbb{E}_{P}\left(\beta\|\tau-\theta\|\right)^{2},\left(\mathbb{E}_{P}\beta\|\tau-\theta\|\right)^{2}\right\} $
		and for $\gamma=\beta$ we can replace $\|\tau-\theta\|\leq\eta-D$
		with $\eta-D$. 
	\end{itemize}
	\item (N-4) $\mathbb{E}_{P}\left[\left|\psi({x};\theta_{0})\right|^{2}\right]<\infty$
	is clear from the analytic expression of $\psi({x};\theta)$, which
	involves at most quadratic entries in ${x}$, and the fact that we assume $P$
	has finite second moments. 
\end{itemize}
Assumptions (N-1) through (N-4) allow us to apply Theorem 3 and its
corollary in \citet{huber1967behavior} and claim
Theorem B. 

The asymptotic normality result allows us to claim
a Wald-type hypothesis testing for the estimated center and radius
for the sphere for a fixed index set $\mathcal{I}$. %We do not pursue
that aspect in the current paper but point out that this is one of
the few non-bootstrap hypothesis testing methods in manifold learning literature.

\subsection{\label{sec:Thm_MSE}Proof of Theorem \ref{thm: MSE comparison}}
First we compare SRCA with PCA. Assume $\|{x}_i\|\leq \alpha$ for any $i$, then for any $\epsilon>0$, there exists a sphere $S_\epsilon$ such that $d({y},S_\epsilon)\leq\epsilon$ for any ${y}\in H$ with $\|{y}\|\leq \alpha$ \citep{li_efficient_2022}. Intuitively, a plane can be approximated by a sphere with infinite radius. Let $\widehat{{x}}_i=\arg\min_{{y}\in H}d({x}_i,{y})$ be the linear projection of ${x}_i$ to plane $H$, then by the triangle inequality,
$$d({x}_i,S_\epsilon)\leq d({x}_i,\widehat{{x}}_i)+d(\widehat{{x}}_i,S_\epsilon).$$
Since the linear projection of a bounded set is still bounded, $d(\widehat{{x}}_i,S_\epsilon)\leq\epsilon$. By the definition of SRCA,
\begin{align*}
	\sum_{i=1}^n d^2({x}_i,S_2)\leq \sum_{i=1}^n d^2({x}_i,S_\epsilon)
	\leq \sum_{i=1}^n d^2({x}_i,H)+2\epsilon  \sum_{i=1}^n d({x}_i,H)+n\epsilon^2.
\end{align*}
Let $\epsilon\to0$, we conclude that 
$$\sum_{i=1}^n d^2({x}_i,S_2)\leq\sum_{i=1}^n d^2({x}_i,H).$$
Then we compare SRCA with SPCA. Since the objective function $\mathscr{L}$, which defines SRCA, is $\min_{S}\sum_{i=1}^n d^2({x}_i,S)$, it follows from $S_1\subset H$ that 
$$\sum_{i=1}^n d^2({x}_i,S_2)\leq \sum_{i=1}^n d^2({x}_i,S_1).$$
Note that SPCA is a restricted version of SRCA, where $\mathcal{I}=\{1,\cdots,d'+1\}$.

\section{\label{sec:out-of-sample MSE}Mean Square Errors for Out-of-sample Data}
This section provides the out-of-sample mean square errors of PCA, SPCA and SRCA on the same datasets in Table \ref{table:MSE_BIG_TABLE}.
We provide this to show that performance evaluation measures are not really affected by the choice of testing samples.

\begin{table}[ht!]
	\centering 
	\global\long\def\~{\hphantom{0}}%
	
	%\begin{turn}{90}
	\begin{tabular}{cccccc}
		Dataset
		& Method/$d'=$ & 1 & 2 & 3 & 4\tabularnewline
		\midrule
		\midrule 
		\multirow{3}{*}{Banknote} & PCA & 15.6094   & 6.2737  &  1.9278 & \tabularnewline
		\cmidrule{2-6} 
		& SPCA & 15.0516   & 7.3182  &  1.5511 &    \tabularnewline
		\cmidrule{2-6}  
		& SRCA & \bf{13.2273} & \bf{5.4120} & \bf{1.1257} &    \tabularnewline
		\midrule
		Power & PCA & 227.6291 &  56.0524 &  23.4302   & \bf{3.0244}  \tabularnewline
		\cmidrule{2-6}  
		Plant & SPCA & 151.7555 & 102.3262  & 44.3802 &  41.0081    \tabularnewline
		\cmidrule{2-6} 
		& SRCA & \bf{151.3426} &  \bf{52.9769} &  \bf{20.0871}  &  4.0775  \tabularnewline
		\midrule 
		\multicolumn{1}{c}{User } & PCA & 0.1952 & 0.1281  & 0.0749  & 0.0306   \tabularnewline
		\cmidrule{2-6}  
		Knowledge & SPCA & 0.1478  & 0.0898  & 0.0465  & 0.0145  \tabularnewline
		\cmidrule{2-6}  
		& SRCA & \bf{0.1479}  & \bf{0.0904}  & \bf{0.0462}  & \bf{0.0144}   \tabularnewline
		\midrule 
		\multirow{3}{*}{Ecoli} & PCA & 0.0761 &   0.0334  &  0.0219  & \bf{ 0.0057}   \tabularnewline
		\cmidrule{2-6} 
		& SPCA & \bf{0.0462} &   0.0351  &  0.0187  &  0.0122    \tabularnewline
		\cmidrule{2-6}  
		& SRCA & 0.0758    & \bf{0.0337}   & \bf{0.0168} &   0.0058 \tabularnewline
		\midrule
		\multirow{3}{*}{Concrete} & PCA & 6.8783 &   4.8345 &   3.5462   & 2.5046 \tabularnewline
		\cmidrule{2-6}  
		& SPCA &\bf{5.5565}  &  \bf{4.2051}  &  \bf{3.1664}  &  \bf{2.0173}  \tabularnewline
		\cmidrule{2-6} 
		& SRCA & 5.5573  &  4.2173  &  3.1842  &  2.0389  \tabularnewline
		\midrule 
		\multirow{3}{*}{Leaf} & PCA & 0.0245 &   0.0126 &   \bf{0.0062}  &  0.0040 \tabularnewline
		\cmidrule{2-6} 
		& SPCA & \bf{0.0163}   & \bf{0.0100} &   0.0073  &  0.0047\tabularnewline
		\cmidrule{2-6} 
		& SRCA & 0.0164  &  0.0101 &  \bf{ 0.0062}  & \bf{0.0037}
		\tabularnewline
		\midrule 
		\multirow{3}{*}{Climate} & PCA & 1.4486  &  1.3846   & 1.3167  &  1.2447  \tabularnewline
		\cmidrule{2-6} 
		& SPCA & 1.4265 &   1.3637  &  1.2921   & 1.2224 \tabularnewline
		\cmidrule{2-6} 
		& SRCA & \bf{1.3863}&   \bf{ 1.3081}  &  \bf{1.2278} &   \bf{1.1525}
		\tabularnewline
		\bottomrule
	\end{tabular}
	\caption{Out-of-sample mean square error (MSE) table for different experiments. }\label{table:OoSMSE_BIG_TABLE}
\end{table}

\end{document}